\documentclass[english]{article}
\usepackage[T1]{fontenc}
\usepackage[latin9]{inputenc}
\usepackage{babel}
\usepackage{array}
\usepackage{float}
\usepackage{multirow}
\usepackage{amsmath}
\usepackage{amssymb}
\usepackage{graphicx}
\usepackage[unicode=true]
 {hyperref}

\makeatletter

\providecommand{\tabularnewline}{\\}

\PassOptionsToPackage{numbers, compress}{natbib}

\usepackage[final]{neurips_2022}

\usepackage[T1]{fontenc}    
\usepackage{hyperref}       
\usepackage{url}            
\usepackage{booktabs}       
\usepackage{amsfonts}       
\usepackage{nicefrac}       
\usepackage{microtype}      
\usepackage{xcolor} 
\usepackage{caption}

\usepackage{pifont}

\@ifundefined{showcaptionsetup}{}{%
 \PassOptionsToPackage{caption=false}{subfig}}
\usepackage{subfig}
\makeatother

\begin{document}

\title{Momentum Adversarial Distillation: Handling Large Distribution Shifts
in Data-Free Knowledge Distillation}

\author{\textbf{Kien Do, Hung Le, Dung Nguyen, Dang Nguyen, Haripriya Harikumar,
}\\
\textbf{Truyen Tran, Santu Rana, Svetha Venkatesh}\\
Applied Artificial Intelligence Institute (A2I2), Deakin University,
Australia\\
\emph{\{k.do, thai.le, dung.nguyen, d.nguyen, h.harikumar, }\\
\emph{truyen.tran, santu.rana, svetha.venkatesh\}@deakin.edu.au}}

\maketitle
\global\long\def\Expect{\mathbb{E}}
\global\long\def\Real{\mathbb{R}}
\global\long\def\Data{\mathcal{D}}
\global\long\def\Loss{\mathcal{L}}
\global\long\def\Normal{\mathcal{N}}
\global\long\def\IdentityMat{\mathrm{I}}
\global\long\def\Tea{\mathtt{T}}
\global\long\def\Teap{\mathtt{Tp}}
\global\long\def\Stu{\mathtt{S}}
\global\long\def\Stup{\mathtt{Sp}}
\global\long\def\Emb{\mathtt{E}}
\global\long\def\Gen{\mathtt{G}}
\global\long\def\Gena{\tilde{\mathtt{G}}}
\global\long\def\argmin#1{\underset{#1}{\text{argmin}}}
\global\long\def\argmax#1{\underset{#1}{\text{argmax}}}
\global\long\def\softmax{\text{softmax}}
\global\long\def\oh{\text{oh}}
\global\long\def\Model{\text{MAD}}

\begin{abstract}
Data-free Knowledge Distillation (DFKD) has attracted attention recently
thanks to its appealing capability of transferring knowledge from
a teacher network to a student network without using training data.
The main idea is to use a generator to synthesize data for training
the student. As the generator gets updated, the distribution of synthetic
data will change. Such distribution shift could be large if the generator
and the student are trained adversarially, causing the student to
forget the knowledge it acquired at previous steps. To alleviate this
problem, we propose a simple yet effective method called Momentum
Adversarial Distillation ($\Model$) which maintains an exponential
moving average (EMA) copy of the generator and uses synthetic samples
from both the generator and the EMA generator to train the student.
Since the EMA generator can be considered as an ensemble of the generator's
old versions and often undergoes a smaller change in updates compared
to the generator, training on its synthetic samples can help the student
recall the past knowledge and prevent the student from adapting too
quickly to new updates of the generator. Our experiments on six benchmark
datasets including big datasets like ImageNet and Places365 demonstrate
the superior performance of $\Model$ over competing methods for handling
the large distribution shift problem. Our method also compares favorably
to existing DFKD methods and even achieves state-of-the-art results
in some cases.

\end{abstract}

\section{Introduction}

With the development of deep learning, more pretrained deep neural
networks have been released to the public \cite{carion2020end,devlin2018bert,he2016deep,radford2021learning,vaswani2017attention}.
However, their superior performances often come with big sizes, causing
difficulties in deployment of these pretrained networks on resource-constrained
devices. This leads to the demand for transferring knowledge from
a cumbersome pretrained source network (called ``teacher'') to a
compact target network (called ``student'') with a minimal loss
of performance. This task is regarded as Knowledge Distillation (KD)
\cite{hinton2015distilling}.

The original idea of KD is to make use of class probabilities predicted
by the teacher which encapsulate the hidden correlation among classes
as training signals for the student. Note that such ``dark'' knowledge
\cite{hinton2015distilling} is generally not available if the student
is trained directly on raw data with one-hot labels. Later, various
KD methods have been proposed to improve the quality of knowledge
transfer. For example, AT \cite{zagoruyko2016paying} matches the
spatial attention maps at intermediate layers of the student and the
teacher. SPD \cite{tung2019similarity} encourages the similarity
between the student's and teacher's feature correlation matrices.
PKT \cite{passalis2018learning} transfers the conditional probability
between two samples computed via kernel density estimation on the
feature space. RKD \cite{park2019relational} exploits relational
knowledge for distillation. VID \cite{ahn2019variational} maximizes
a variational lower bound of the mutual information (MI) between the
student's and teacher's representations. CRD \cite{tian2019contrastive}
uses contrastive learning as a proxy for maximizing MI.

The common drawback of these methods is the reliance on samples from
the teacher training set. However, in practice, accessing the original
training data is usually infeasible due to many reasons such as data
privacy (e.g., healthcare data containing personal information) or
data regarded as intellectual property of the vendors. Addressing
this critical issue, Data-Free Knowledge Distillation (DFKD) methods
have been introduced \cite{chen2019data,fang2019data,lopes2017data,luo2020large,micaelli2019zero,nayak2019zero,wang2021data,yin2020dreaming,yoo2019knowledge}.
A common DFKD approach is to use a generator network to synthesize
training data and jointly train the generator and the student in an
adversarial manner \cite{fang2019data,micaelli2019zero,yoo2019knowledge}.
Under this adversarial learning scheme, the student attempts to make
predictions as close as possible to the teacher's on synthetic data
generated by the generator, while the generator tries to create samples
that maximize the mismatch between the student's and the teacher's
predictions. This adversarial game enables a rapid exploration of
synthetic distributions useful for knowledge transfer between the
teacher and the student. At the same time, it could also lead to large
shifts in the synthetic distributions, causing the student to forget
useful knowledge acquired at the previous steps and suffer from performance
drops \cite{binici2022preventing}.

In this paper, we propose a simple yet effective method called Momentum
Adversarial Distillation ($\Model$) to mitigate the large distribution
shift problem in adversarial DFKD. $\Model$ maintains an exponential
moving average (EMA) copy of the generator which is responsible for
storing information about past updates of the generator. By using
synthetic samples from the EMA generator as additional training data
for the student besides those from the generator, $\Model$ can ensure
that the student can recall the old knowledge, hence, is less prone
to forgetting. Moreover, to reduce the negative effect caused by spurious
solutions of an unconditional generator when learning on large datasets
such as ImageNet, we propose to use a class-conditional generator
that takes the sum of a noise vector and a class embedding vector
as input, and train this generator with a new objective that suppresses
the presence of spurious solutions. This technique requires only a
small change in the generator's architecture but enables $\Model$
(and possibly other adversarial DFKD methods) to learn surprisingly
well on large datasets. Through extensive experiments on three small
and three large image datasets, we demonstrate that our proposed method
is far better than related baselines \cite{binici2022preventing,micaelli2019zero}
in dealing with the large distribution shift problem. In some cases,
$\Model$ even outperforms current state-of-the-art methods \cite{choi2020data,fang2021contrastive}.

\section{Adversarial Data-Free Knowledge Distillation\label{sec:Prelim}}

Let $\Tea$ be a \emph{teacher} network pretrained on some dataset
$\Data_{\text{train}}$ and $\Stu$ be a fresh \emph{student} network.
Let $\Tea(\cdot)$ and $\Stu(\cdot)$ denote outputs of the teacher
and student networks \emph{before the softmax activation}, respectively.
In Data-Free Knowledge Distillation (DFKD), we want to transfer knowledge
from $\Tea$ to $\Stu$ so that $\Stu$ performs as well as or even
better than $\Tea$ on the original test set $\Data_{\text{test}}$
but with a constraint that no training data for $\Stu$ is available.
An intuitive way to deal with such constraint is learning an additional
generator network $\Gen$ that can generate synthetic data for training
$\Stu$ from a noise distribution $p(z)$ usually chosen to be the
standard Gaussian distribution $\Normal(0,\IdentityMat)$. Adversarial
Belief Matching (ABM) \cite{micaelli2019zero} proposed an adversarial
learning framework between $\Stu$ and $\Gen$ via optimizing the
following min-max objective:
\begin{align}
 & \min_{\Stu}\max_{\Gen}\Expect_{z\sim p(z)}\left[\Loss_{\text{KD}}(\Gen(z))\right]\label{eq:minimax_objective}\\
\Leftrightarrow & \min_{\Stu}\max_{\Gen}\Expect_{z\sim p(z),x=\Gen(z)}\left[\Loss_{\text{KD}}(x)\right],\label{eq:minimax_objective_1}
\end{align}
where $\Loss_{\text{KD}}(x)$ denotes the knowledge distillation (KD)
loss, i.e., the discrepancy between $\Stu(x)$ and $\Tea(x)$. In
ABM, $\Loss_{\text{KD}}(x)$ is the Kullback-Leibler (KL) divergence
between class probabilities of $\Tea$ and $\Stu$ computed on $x$:
\begin{align}
\Loss_{\text{KD}}(x) & \triangleq D_{\text{KL}}\left(\Teap(x)\|\Stup(x)\right)=\sum_{c=1}^{C}\Teap(x)[c]\cdot\left(\log\Teap(x)[c]-\log\Stup(x)[c]\right),\label{eq:KDLoss_as_KL}
\end{align}
where $\Teap(x)=\softmax(\Tea(x))$ and $\Stup(x)=\softmax(\Stu(x))$
denote the class probabilities of $\Tea$ and $\Stu$ computed on
$x$, respectively; and $C$ is the total number of classes.

The core idea behind the optimization in Eq.~\ref{eq:minimax_objective}
is to encourage $\Gen$ to generate samples on which the outputs of
$\Stu$ are very different from those of $\Tea$ (or the KD loss is
large). Typically, the generated samples have not been observed by
$\Stu$ during training (otherwise, the KD loss will be small) and
the learning task of $\Stu$ is to match $\Tea$ on these novel samples.
It is expected that in the ideal case, via continuous adversarial
exploration of $\Gen$, $\Stu$ can be exposed to a diverse enough
set of synthetic samples which allows $\Stu$ to match $\Tea$ with
an arbitrarily small prediction error on $\Data_{\text{test}}$.

In practice, we usually implement Eq.~\ref{eq:minimax_objective}
by alternately optimizing $\Stu$ and $\Gen$ in $n_{\Stu}$ and $n_{\Gen}$
steps, respectively. Thus, to avoid confusion later, we refer to a
$n_{\Stu}$-step update of $\Stu$ as an update \emph{stage} of $\Stu$
and similarly, a $n_{\Gen}$-step update of $\Gen$ as an update \emph{stage}
of $\Gen$.

\section{Momentum Adversarial Distillation}

\begin{figure}
\begin{centering}
\includegraphics[width=0.75\textwidth]{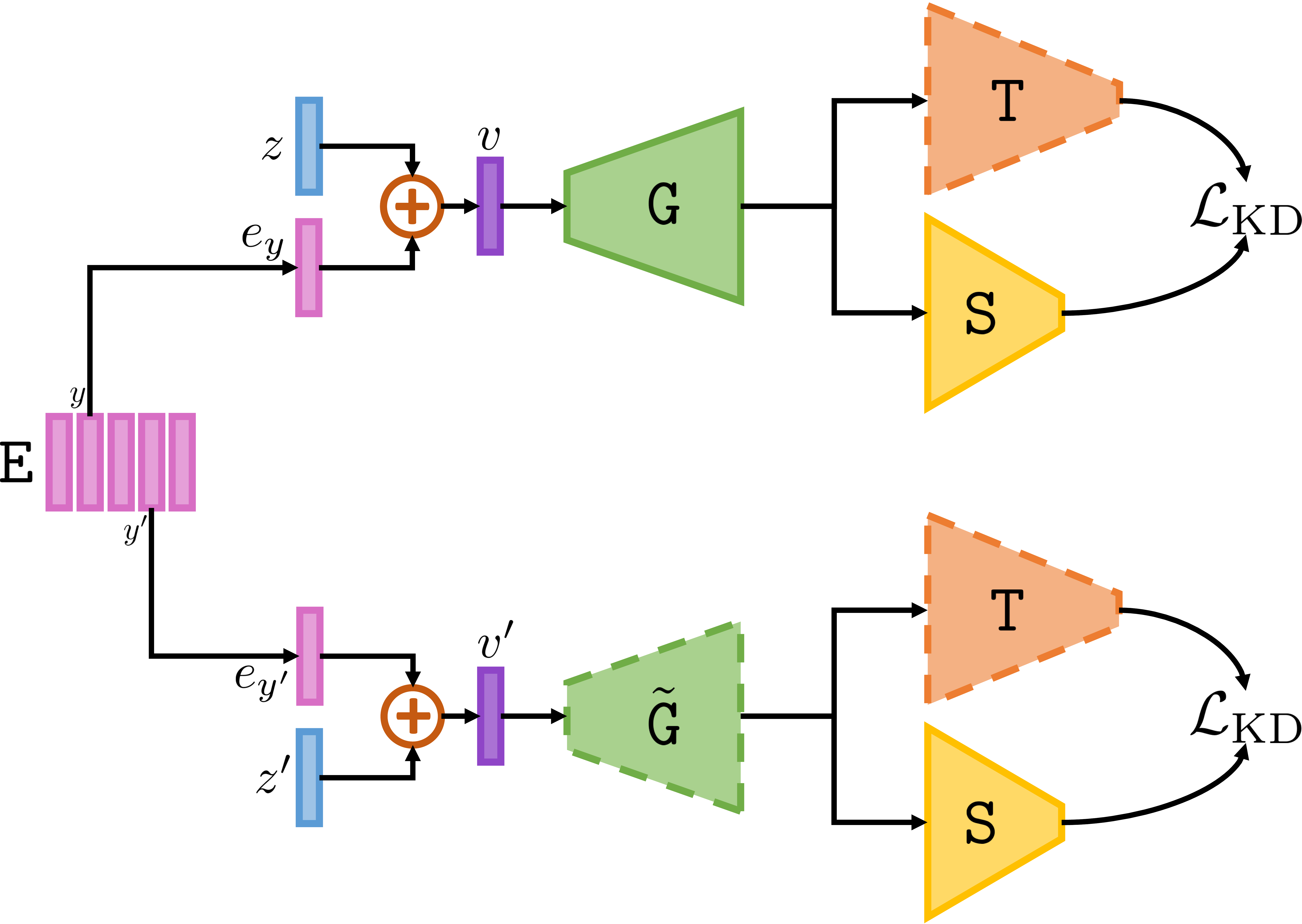}
\par\end{centering}
\caption{An illustration of our proposed \emph{Momentum Adversarial Distillation
($\protect\Model$)} consisting of a teacher ($\protect\Tea$), a
student ($\protect\Stu$), a class-conditional generator ($\protect\Gen$),
an EMA generator ($\tilde{\protect\Gen}$), and class embeddings ($\protect\Emb$).
Networks with dashed borders ($\protect\Tea$, $\tilde{\protect\Gen}$)
are not optimized during training. $z$ and $z'$ are random noises
sampled from $\protect\Normal(0,\mathrm{I})$. $e_{y}$ and $e_{y'}$
are gathered from $\protect\Emb$ at index $y$ and $y'$, respectively.\label{fig:MAD_diagram}}
\end{figure}

\subsection{Handling large distribution shifts with an additional EMA generator }

Although theoretically sound, the optimization in Eq.~\ref{eq:minimax_objective}
has a practical problem: If the update of $\Gen$ varies too much,
the distribution of synthetic samples generated by $\Gen$ will change
significantly over two consecutive update steps of $\Stu$, which
in turns causes $\Stu$ to catastrophically forget what it has learned
at the previous stages \cite{binici2022preventing} to adapt to the
new update of $\Gen$. In order to address this problem, we propose
to maintain an exponential moving average (EMA) of the generator $\Gen$,
denoted by $\tilde{\Gen}$, during learning and use synthetic samples
from both $\Gen$ and $\tilde{\Gen}$ to train $\Stu$ as follows:
\begin{equation}
\min_{\Stu}\Loss_{\Stu}\triangleq\lambda_{0}\Expect_{z\sim p(z)}\left[\Loss_{\text{KD}}(\Gen(z))\right]+\lambda_{1}\Expect_{z'\sim p(z)}\left[\Loss_{\text{KD}}(\tilde{\Gen}(z'))\right],\label{eq:MAD_LossS}
\end{equation}
where $\lambda_{0},\lambda_{1}\geq0$ are coefficients. The parameters
$\theta_{\tilde{\Gen}}^{t}$ of the momentum generator $\tilde{\Gen}$
at the update \emph{stage} $t$ of $\Stu$ (after $n_{\Gen}$ steps
update of $\Gen$) are computed as:
\begin{equation}
\theta_{\tilde{\Gen}}^{t}=\alpha\cdot\theta_{\tilde{\Gen}}^{t-1}+(1-\alpha)\cdot\theta_{\Gen}^{t},\label{eq:momentum}
\end{equation}
where $\alpha$ ($0<\alpha<1$) is the momentum. If $\alpha$ is close
to 1, $\tilde{\Gen}$ will change very slightly compared to $\Gen$.
Therefore, by using synthetic samples from $\tilde{\Gen}$ as additional
training data for $\Stu$ besides those from $\Gen$, we can alleviate
the large exploratory distribution shift caused by the large update
of $\Gen$ and achieve a stable update of $\Stu$. We name our proposed
method \emph{Momentum Adversarial Distillation} ($\Model$). See Fig.~\ref{fig:MAD_diagram}
for an illustration.

\subsection{Enabling $\protect\Model$ to learn well on large datasets\label{subsec:ClassConditionalGen}}

During our experiments, we observed that $\Model$, with the unconditional
generator $\Gen$ described in Section~\ref{sec:Prelim}, is not
able to learn well on large datasets such as ImageNet. Our hypothesis
is that in case of large datasets, the objective of $\Gen$ in Eq.~\ref{eq:minimax_objective_1}
induces a very large number spurious solutions, which hampers the
learning of $\Gen$. To see this, first let us recall the objective
of $\Gen$ which is maximizing the KL divergence between $\Stup(x)$
and $\Teap(x)$ over some synthetic sample $x=\Gen(z)$. If we, for
example, assume that there are 3 classes in total and $\Teap(x)$
is fixed at $[1,0,0]$, then we could train $\Gen$ to generate $x$
so that $\Stup(x)$ is either $[0,1,0]$ or $[0,0,1]$. In this toy
example, we see 2 spurious solutions for 3 classes. If a dataset has
1000 classes like ImageNet, there will be at least 999 spurious solutions.
Besides class numbers, larger input sizes also increase the space
and the number of (spurious) solutions. Unfortunately, the more spurious
solutions, the more likely $\Gen$ could jump from one spurious solution
to another in successive update steps, causing instability in training
$\Gen$.

To overcome this problem, we propose to condition $\Gen$ on a class
label $y$ as $\Gen(z+e_{y})$ where $e_{y}$ is a trainable embedding
of $y$. We found that using the sum of $z$ and $e_{y}$ as input
to $\Gen$ instead of concatenation allows our method to learn much
better, possibly because the noise in updating $e_{y}$ is absorbed
into the stochasticity of $z$ via summation rather than concatenation.
We illustrate this idea in Fig.~\ref{fig:MAD_diagram}, and provide
an empirical justification in Appdx.~\ref{subsec:Results-Diff-Cond-Types}.
Denoted by $\Emb$ the list of trainable embedding vectors for all
classes ($\Emb=(e_{1},...,e_{C})$), we train both $\Gen$ and $\Emb$
together by minimizing the following loss: 
\begin{equation}
\min_{\Gen,\Emb}\Loss_{\Gen,\Emb}\triangleq\Expect_{z\sim\mathcal{N}(0,\mathrm{I}),y\sim\text{Cat}(C),x=\Gen(z+e_{y})}\left[-\lambda_{2}\Loss_{\text{KD}}(x)+\lambda_{3}\Loss_{\text{NLL}}(x,y)+\lambda_{4}\Loss_{\text{NormReg}}(e_{y})\right],\label{eq:ClassCondGen_loss}
\end{equation}
where $\text{Cat}(C)$ is the uniform categorical distribution over
$C$ classes; $\lambda_{2},\lambda_{3},\lambda_{4}\geq0$ are hyperparameters;
and $\Loss_{\text{NLL}}(x,y)$ and $\Loss_{\text{NormReg}}(e_{y})$
are the \emph{negative log-likelihood} and the \emph{norm regularization}
losses respectively, defined as follows:
\begin{align}
\Loss_{\text{NLL}}(x,y) & \triangleq-\log\Teap(x)[y],\label{eq:NLL_loss}\\
\Loss_{\text{NormReg}}(e_{y}) & \triangleq\max\left({\left\Vert e_{y}\right\Vert }_{2}-\gamma\times\sqrt{d_{e}},0\right),\label{eq:NormReg_loss}
\end{align}
where $d_{e}$ denotes the dimensionality of $e_{y}$; and $\gamma\geq1$
is a scaling hyperparameter. $\Loss_{\text{NormReg}}(e_{y})$ servers
as a constraint that restricts the norm of $e_{y}$ to be smaller
than $\gamma\times\sqrt{d_{e}}$. An explanation of the formula $\gamma\times\sqrt{d_{e}}$
is provided in Appdx.~\ref{subsec:Derivation-of-NormRegLoss}. Intuitively,
via minimizing $\Loss_{\text{NLL}}(x,y)$, $\Gen$ could maintain
its focus on predicting $y$ throughout its entire update stage rather
than jumping between different spurious solutions. Besides, since
$y$ is sampled uniformly, the synthetic data will be evenly distributed
among classes.

In case the teacher $\Tea$ has BatchNorm \cite{ioffe2015batch} layers,
we can make use of the running mean $\bar{\mu}_{\ell}$ and running
variance $\bar{\omega}_{\ell}$ of each BatchNorm layer $\ell$ of
$\Tea$ to guide the data synthesis of $\Gen$ by adding the BatchNorm
moment matching (BNmm) loss \cite{yin2020dreaming} below to $\Loss_{\Gen,\Emb}$
(weighted by $\lambda_{5}\geq0$):

\begin{equation}
\Loss_{\text{BNmm}}\triangleq\sum_{\ell}\left\Vert \mu_{\ell}-\bar{\mu}_{\ell}\right\Vert _{2}^{2}+\left\Vert \omega_{\ell}-\bar{\omega}_{\ell}\right\Vert _{2}^{2},\label{eq:BNmm_loss}
\end{equation}
where $\mu_{\ell}$, $\omega_{\ell}$ are the empirical mean and variance
of the features at the BatchNorm layer $\ell$ w.r.t. the batch of
synthetic samples from $\Gen$. During our experiment, we observed
that using $\Loss_{\text{BNmm}}$ improves knowledge distillation
of the student $\Stu$.

\section{Related Work}

In Data-Free Knowledge Distillation (DFKD), it is critical to synthesize
data that are useful for transferring knowledge from $\Tea$ to $\Stu$.
Existing DFKD methods differ mainly in their objectives to guide the
data synthesis. These methods generally fall into either the adversarial
camp or the non-adversarial camp. Non-adversarial DFKD methods \cite{chen2019data,lopes2017data,luo2020large,nayak2019zero,wang2021data,yoo2019knowledge}
make use of certain heuristics to search for synthetic data that resembles
the original training data $\Data_{\Tea,\text{train}}$. For example,
ZSKD \cite{nayak2019zero} and DAFL \cite{chen2019data} consider
prediction probabilities of $\Tea$ and the confidence of $\Tea$
as heuristics. In DAFL, synthetic data is generated via a generator
$\Gen$ rather than being optimized directly like in ZSKD. KegNet
\cite{yoo2019knowledge} extends DAFL by making $\Gen$ conditioned
on class labels. Adversarial DFKD methods \cite{choi2020data,fang2021contrastive,han2021robustness,qu2021enhancing,yin2020dreaming,zhao2021dual}
leverage adversarial learning to explore the data space more efficiently.
Most methods of this type are derived from the ABM \cite{micaelli2019zero}
discussed in Section~\ref{sec:Prelim} with additional objectives
to improve the quality and/or diversity of synthetic data. For example,
RDSKD \cite{han2021robustness} uses a diversity-seeking loss \cite{mao2019mode};
CMI \cite{fang2021contrastive} uses an inverse contrastive loss to
improve diversity; and DeepInversion \cite{yin2020dreaming} uses
Batch Norm moment matching (BNmm) and DeepDream's inception losses
(total variation, L2) \cite{mordvintsev2015inceptionism} to generate
visually interpretable images.

DeepInversion \cite{yin2020dreaming} is one of a few DFKD methods
\cite{fang2021up,luo2020large,yin2020dreaming} that have been shown
to learn well on the full-size ImageNet dataset. Its success has inspired
subsequent works on large-scale continual learning \cite{smith2021always}
and data-free object detection \cite{chawla2021data}. However, training
DeepInversion on ImageNet is very time-consuming because this method
optimizes synthetic images directly and \emph{each} complete optimization
requires a very large number of iterations (about 20,000). LS-GDFD
\cite{luo2020large} learns to generate synthetic data instead but
requires one generator for each class (that is, 1,000 generators for
1,000 classes) to avoid the ``mode collapse'' problem and achieve
good performance. FastDFKD \cite{fang2021up} leverages meta-learning
to speed up the training process significantly. On the contrary, by
applying our proposed technique in Section~\ref{subsec:ClassConditionalGen},
our $\Model$ only needs 1 generator (compared to 1,000 of LS-GDFD)
and 6,000 training iterations (compared to 20,000$\times$ of DeepInversion)
to achieve reasonably good performance on ImageNet (Section~\ref{subsec:Comparison-with-related-baselines}).
In addition, our technique is orthogonal to and can combine with the
meta-learning idea in \cite{fang2021up} for further improvement.

Besides the DFKD methods discussed above, there is a line of works
that consider a less extreme scenario in which unlabeled transfer
sets are given for training. These transfer sets can be very different
from the original dataset in terms of distribution and semantics and
can be freely collected from open resources. Nayak et al. \cite{nayak2021effectiveness}
analyzed various kinds of transfer sets ranging from random noise
images to natural images and found that the target-class balance property
of the transfer set and the similarity between the transfer set and
the original training set play important roles in improving knowledge
distillation results. Fang et al. \cite{fang2021mosaicking} argued
that although the transfer images can be semantically different from
the original traning images, they still share common local visual
patterns. Therefore, the authors proposed an interesting method called
MosaicKD which combines different local patches extracted from the
transfer images to craft synthetic mosaic images that capture the
semantics of the original data while enjoy realistic local structures.

Adversarial training methods like GANs has been shown to suffer from
the catastrophic forgetting problem of the discriminator \cite{liang2018generative,thanh2020catastrophic}.
These works consider GANs training as a continual learning problem
and leverage existing methods to mitigate the issue. For example,
Liang et al. \cite{liang2018generative} use either EWC \cite{kirkpatrick2017overcoming}
or SI \cite{zenke2017continual} to enforce similarity between the
discriminator's weights at the current step and at the last checkpoint.
Hoang et al. \cite{thanh2020catastrophic} suggest some other tricks
such as using a momentum optimizer (e.g., Adam \cite{Kingma_Ba_14Adam})
or applying gradient penalty \cite{gulrajani2017improved} to the
discriminator. In this paper, we consider the catastrophic forgetting
problem of the student as a consequence of the large distribution
shift problem caused by the generator. Therefore, we focus on regularizing
the generator rather than the student. Besides, from this perspective,
the catastrophic forgetting problem in GANs is indeed less severe
than that in adversarial DFKD. It is because in GANs, the discriminator
is trained on real data with a fixed distribution, which can somewhat
reduce the effect of the distribution shift of fake (synthetic) data
while in adversarial DFKD, real data is not available. Recent works
that also address the student forgetting problem in DFKD like ours
include DFKD-Mem \cite{binici2022preventing} and PRE-DFKD \cite{binici2022robust}.
DFKD-Mem \cite{binici2022preventing} stores the past synthetic samples
in a memory bank and uses these samples as additional training data
for the student. Our method, on the other hand, uses an EMA generator
to generate old synthetic samples on-the-fly, which is more memory
efficient and adapts better to the student update. PRE-DFKD \cite{binici2022robust}
models past synthetic data via a VAE \cite{kingma2013auto} and treats
the decoder of this VAE as a replay generator. However, training a
VAE on a continuous stream of synthetic samples could be unstable
and could lead to another catastrophic forgetting problem on its own.
In addition, VAE is not effective for generating images with large
size (e.g., ImageNet images) due to the ``posterior collapse'' problem
\cite{lucas2019understanding}.

\section{Experiments}

\subsection{Experimental Setup}

\paragraph{Datasets}

We consider the image classification task and evaluate our proposed
method on 3 small image datasets (CIFAR10 \cite{krizhevsky2009learning},
CIFAR100 \cite{krizhevsky2009learning}, TinyImageNet \cite{le2015tiny}),
and 3 large image datasets (ImageNet \cite{deng2009imagenet}, Places365
\cite{zhou2017places}, Food101 \cite{bossard14}). Details are provided
in Appdx.~\ref{subsec:Datasets}.

\paragraph{Network architectures}

For the small datasets, we follow \cite{choi2020data,fang2021contrastive}
and consider ResNet34/ResNet18 \cite{he2016deep} and WRN40-2/WRN16-2
\cite{zagoruyko2016wide} for the teacher/student. We observed that
knowledge distillation with the WRN architectures is more challenging
than with the ResNet architectures, possibly because WRN40-2/WRN16-2
have much fewer parameters than ResNet34/ResNet18. For the large datasets,
we use the AlexNet architecture for both the teacher and student for
fast training. Since AlexNet does not have any BatchNorm layer, we
exclude the BNmm loss (Eq.~\ref{eq:BNmm_loss}) from the total loss
of the generator in our experiments on the large datasets. Architectures
of the generator w.r.t. different datasets are given in Appdx.~\ref{subsec:Generator-Architectures}. 

\paragraph{Training settings of teacher }

For ImageNet, we make use of pretrained networks provided by PyTorch.
For other datasets, we train the teacher from scratch. Detailed training
settings of the teacher are given in Appdx.~\ref{subsec:Teacher-Training-Settings}.

\paragraph{Training settings of $\protect\Model$}

If not otherwise specified, we set the momentum $\alpha$ in Eq.~\ref{eq:momentum}
to $0.95$ and the length of the noise vector to 256. We train the
student $\Stu$ using SGD and Adam for the small and large datasets,
respectively. We train the generator $\Gen$ using Adam for both the
small and large datasets. To reduce the difficulty of training $\Gen$
for the large datasets, we pretrain $\Gen$ for some steps before
the main KD training by setting the coefficient of $\Loss_{\text{KD}}$
to 0 and only optimizing the remaining losses in Eq.~\ref{eq:ClassCondGen_loss}.
The generator $\Gen$ is class-conditional (as described in Section~\ref{subsec:ClassConditionalGen})
for the large datasets, and unconditional for the small datasets.
We empirically found that this leads to better results. For further
details about the training settings of $\Model$, please refer to
Appdx.~\ref{subsec:Training-Settings-of-Our-Method}.

\subsection{Results}

\subsubsection{Comparison with existing DFKD methods}

In Table~\ref{tab:Results-Existing-Baselines}, we compare $\Model$
with existing DFKD methods \cite{chen2019data,choi2020data,fang2019data,fang2021contrastive,nayak2019zero,yin2020dreaming,zhao2021dual}
on the small datasets. The results of baselines are taken from \cite{fang2021contrastive,zhao2021dual}.
We checked the results of the teacher trained by us and found that
they are quite similar to the results of the teacher reported in \cite{fang2021contrastive}
and in Table~\ref{tab:Results-Existing-Baselines} (details in Appdx.~\ref{subsec:Results-of-teacher}).
This means our results of $\Model$ are comparable with those of the
baselines. We see that $\Model$ outperforms all the DFKD methods
on CIFAR10 and CIFAR100 in case the teacher and student are ResNets
\cite{he2016deep} ($\heartsuit$). In case the teacher and student
are WideResNets \cite{zagoruyko2016wide} ($\diamondsuit$), our method
still achieves significantly better results than most of the baselines
such as DAFL \cite{chen2019data} and ZSKT \cite{nayak2019zero} and
only performs worse than CMI \cite{fang2021contrastive}. We hypothesize
it is mainly because of the differences in model design and training
settings of CMI and $\Model$. For example, CMI has an additional
contrastive-loss-based objective that encourages the generator's diversity
while $\Model$ does not. CMI also performs more updates of the student
and generator per training step than our method\footnote{We could not find the supplementary material containing the training
settings of CMI. However, from the \href{https://github.com/zju-vipa/CMI}{official code provided by the authors},
we saw that they set $n_{\Gen}=200$ and $n_{\Stu}=2000$ for WideResNet
teacher/student on CIFAR10 while we set $n_{\Gen}=3$ and $n_{\Stu}=60$
(see Appdx.~\ref{subsec:Training-Settings-of-Our-Method}). In general,
setting larger $n_{\Gen}$ and $n_{\Stu}$ often leads to better results
(Section~\ref{subsec:Ablation-Study}) but will increase the training
time.}.

However, our ultimate goal for this section is \emph{not} to show
that $\Model$ can achieve state-of-the-art results in all cases (which
often requires intensive hyper-parameter tuning) but to verify the
soundness of our implementation of $\Model$. In order to see clearly
the advantage of $\Model$, we need to compare $\Model$ with its
related baselines \emph{under the same settings}. This will be presented
in Section~\ref{subsec:Comparison-with-related-baselines}.

\begin{table}
\begin{centering}
\resizebox{\textwidth}{!}{%
\par\end{centering}
\begin{centering}
\begin{tabular}{c|c|cc|ccccccc|c}
Dataset & Arch. & Tea.$^{a}$ & Stu.$^{a}$ & DAFL$^{a}$ & ZSKT$^{a}$ & ADI$^{a}$ & DFAD$^{b}$ & DDAD$^{b}$ & DFQ$^{a}$ & CMI$^{a}$ & $\Model$\tabularnewline
\hline 
\hline 
\multirow{2}{*}{CIFAR10} & $\heartsuit$ & 95.70 & 95.20 & 92.22 & 93.32 & 93.26 & 93.30 & 94.81 & 94.61 & \emph{94.84} & \textbf{94.90}\tabularnewline
 & $\diamondsuit$ & 94.87 & 93.95 & 81.55 & 89.66 & 89.72 & - & - & 92.01 & \emph{92.52} & \textbf{92.64}\tabularnewline
\hline 
\multirow{2}{*}{CIFAR100} & $\heartsuit$ & 78.05 & 77.10 & 74.47 & 67.74 & 61.32 & 69.43 & 75.04 & 77.01 & 77.04 & \textbf{77.31}\tabularnewline
 & $\diamondsuit$ & 75.83 & 73.56 & 40.00 & 28.44 & 61.34 & - & - & 59.01 & \textbf{68.75} & \emph{64.05}\tabularnewline
\hline 
TinyIN & $\heartsuit$ & 66.44 & 64.87 & - & - & - & - & - & \emph{63.73} & \textbf{64.01} & 62.32\tabularnewline
\end{tabular}}
\par\end{centering}
\caption{Classification accuracy (in \%) of the student trained by different
DFKD methods on 3 small image datasets. The teacher/student architecture
settings are ResNet34/ResNet18 ($\heartsuit$) and WRN40-2/WRN16-2
($\diamondsuit$). $^{a}$ and $^{b}$ denote results taken from \cite{fang2021contrastive}
and \cite{zhao2021dual}, respectively. Tea. and Stu. denote the teacher
and student trained from scratch on $\protect\Data_{\text{train}}$.
The best and second best results are highlighted in \textbf{bold}
and \emph{italic}, respectively.\label{tab:Results-Existing-Baselines}}
\end{table}

\subsubsection{Comparison with related baselines\label{subsec:Comparison-with-related-baselines}}

We consider two related baselines of $\Model$ which are ABM \cite{micaelli2019zero}
and DFKD-Mem \cite{binici2022preventing}. ABM learns the student
$\Stu$ with only synthetic samples from $\Gen$. DFKD-Mem, on the
other hand, stores past synthetic samples in a memory bank, and uses
samples from this memory bank (dubbed ``memory samples'') as well
as those generated by $\Gen$ as training data for $\Stu$. We trained
ABM and DFKD-Mem using exactly the same settings for training $\Model$.
For DFKD-Mem, we set the memory size to 8,192. For other memory sizes,
the results remain relatively similar as shown in Appdx.~\ref{subsec:Results-of-DFKD-Mem}.

\begin{figure}
\begin{centering}
\subfloat[CIFAR100]{\begin{centering}
\includegraphics[width=0.22\textwidth]{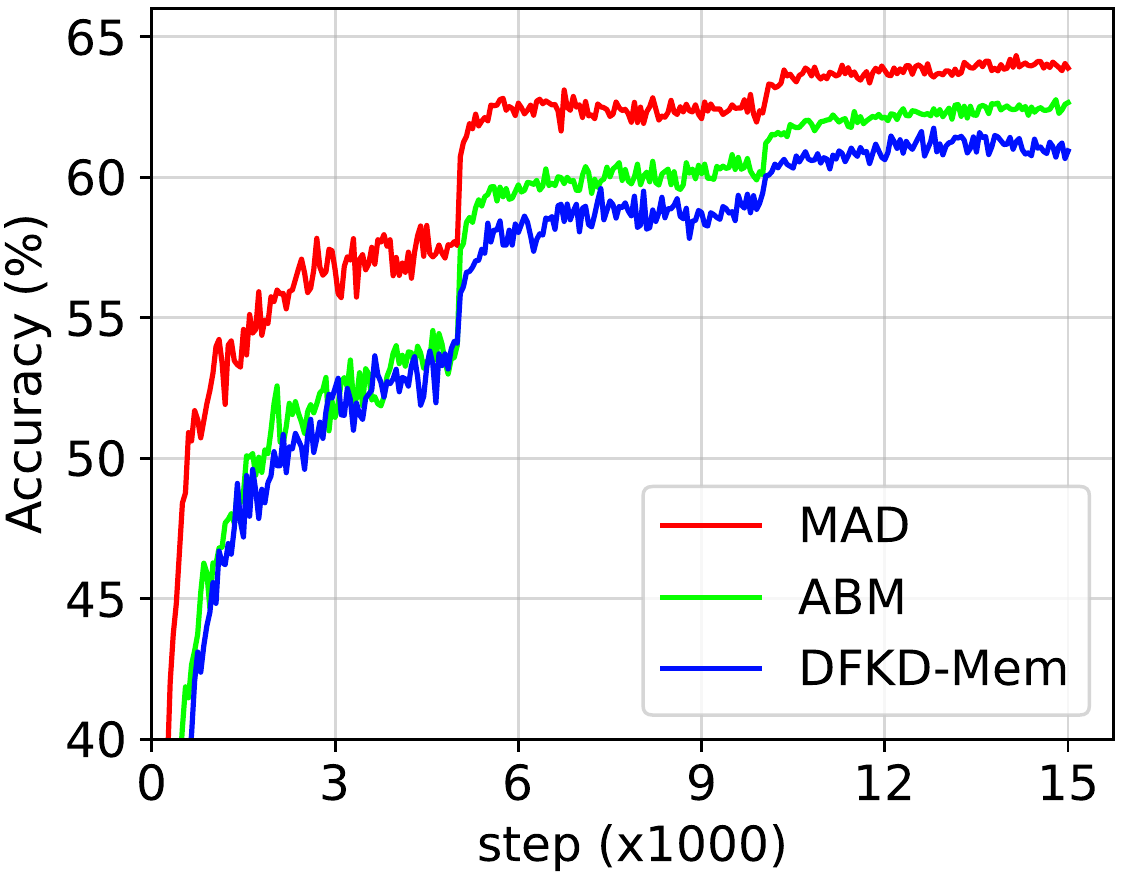}
\par\end{centering}
}\hspace{0.005\textwidth}\subfloat[TinyImageNet]{\begin{centering}
\includegraphics[width=0.22\textwidth]{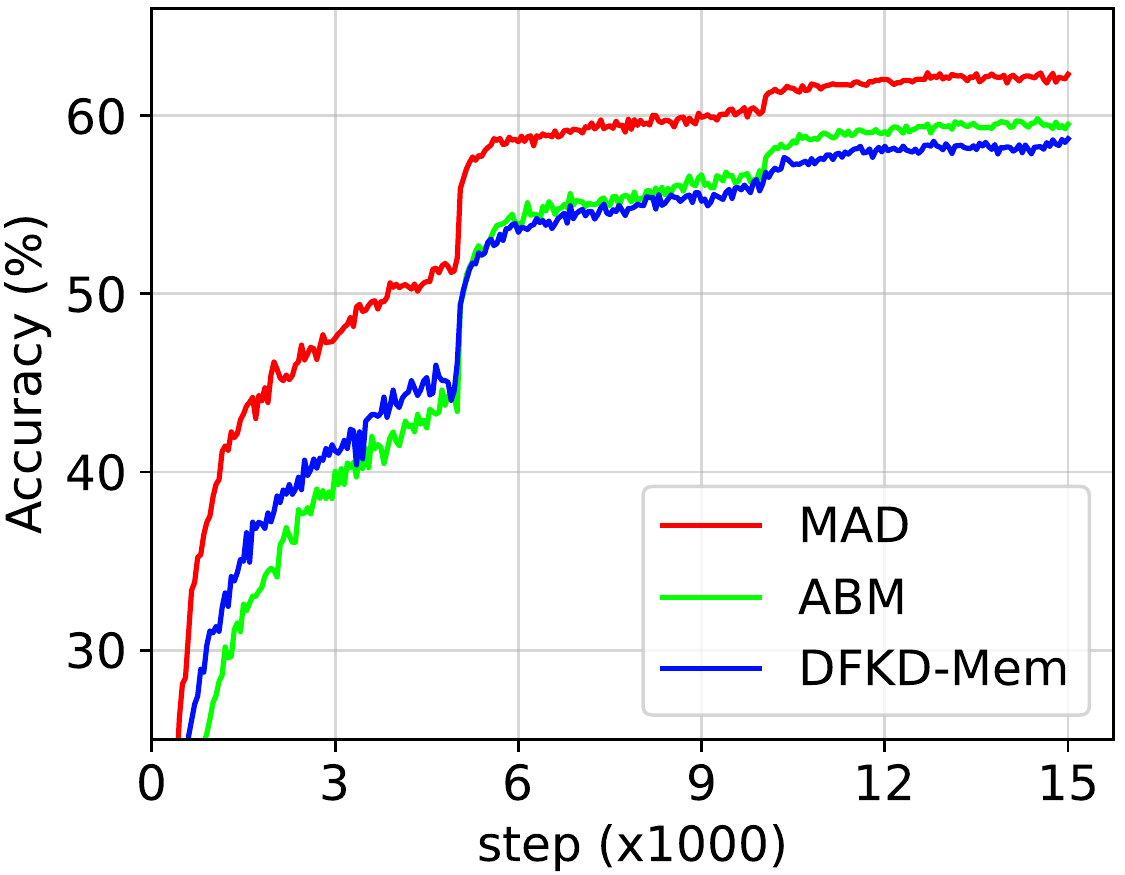}
\par\end{centering}
}\hspace{0.005\textwidth}\subfloat[ImageNet]{\begin{centering}
\includegraphics[width=0.22\textwidth]{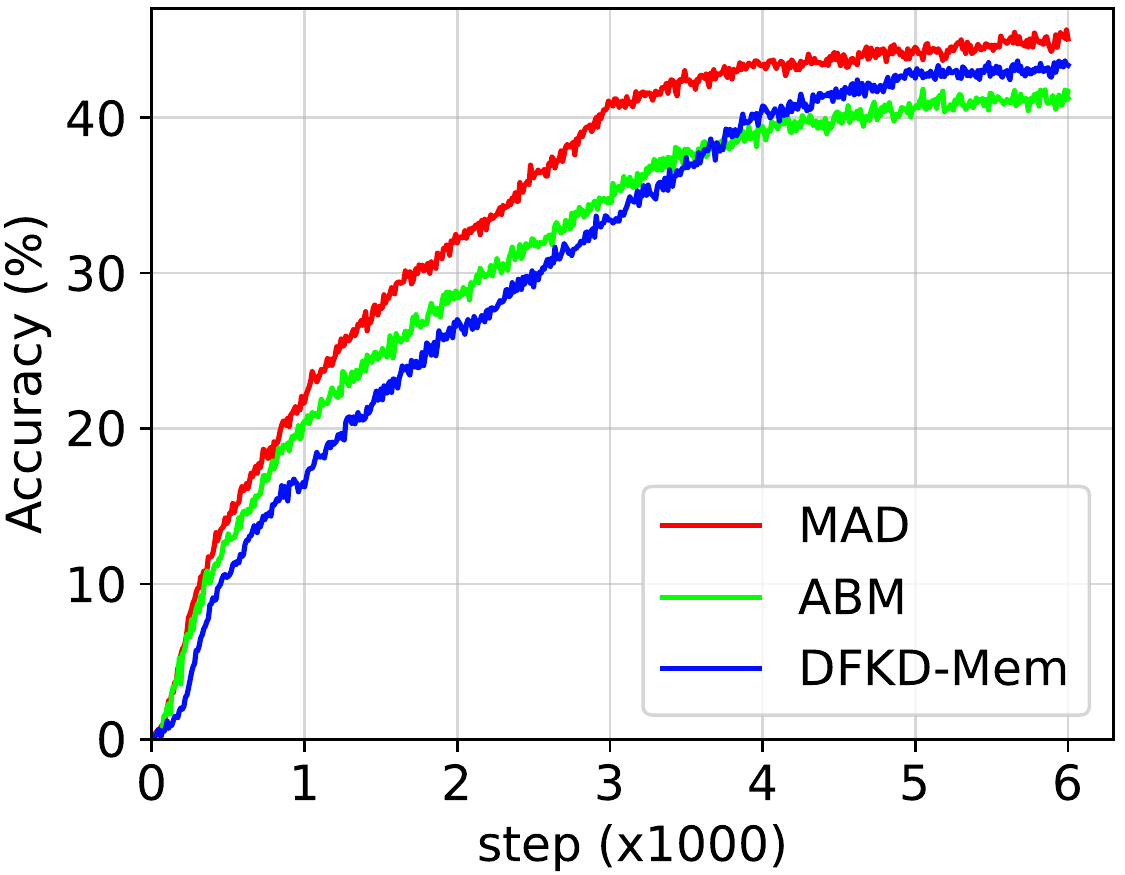}
\par\end{centering}
}\hspace{0.005\textwidth}\subfloat[Places365]{\begin{centering}
\includegraphics[width=0.22\textwidth]{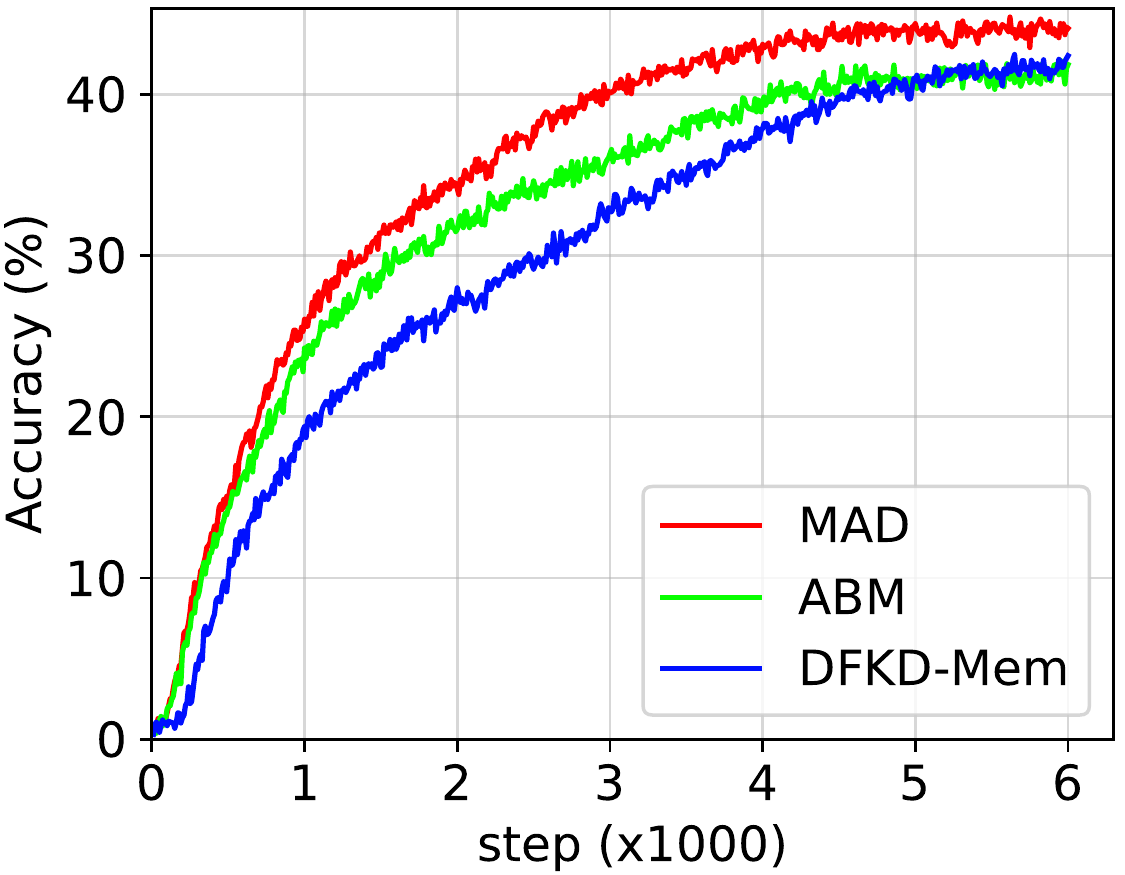}
\par\end{centering}
}
\par\end{centering}
\caption{Test accuracy curves of $\protect\Stu$ trained via $\protect\Model$,
ABM, and DFKD-Mem on some datasets.\label{fig:Test-Acc-Curves-Stu}}
\end{figure}

From Table~\ref{tab:Results-Related-Baselines} and Fig.~\ref{fig:Test-Acc-Curves-Stu},
it is clear that $\Model$ significantly outperforms both ABM and
DFKD-Mem on all datasets. In addition, the performance gaps between
our method and the two baselines tend to be larger for larger datasets.
For example, $\Model$ achieves about 1.5/2.8\%, 2.5/3.6\%, and 4.2/2.2\%
higher accuracy than ABM/DFKD-Mem on CIFAR100, TinyImageNet, and ImageNet,
respectively. These empirical results suggest the importance of the
EMA generator $\tilde{\Gen}$ in mitigating the large distribution
shift caused by $\Gen$. 

\begin{figure}
\begin{centering}
\subfloat[$\protect\Loss_{\text{KD}}$ on memory samples\label{fig:KDLoss_on_que}]{\begin{centering}
\includegraphics[width=0.3\textwidth]{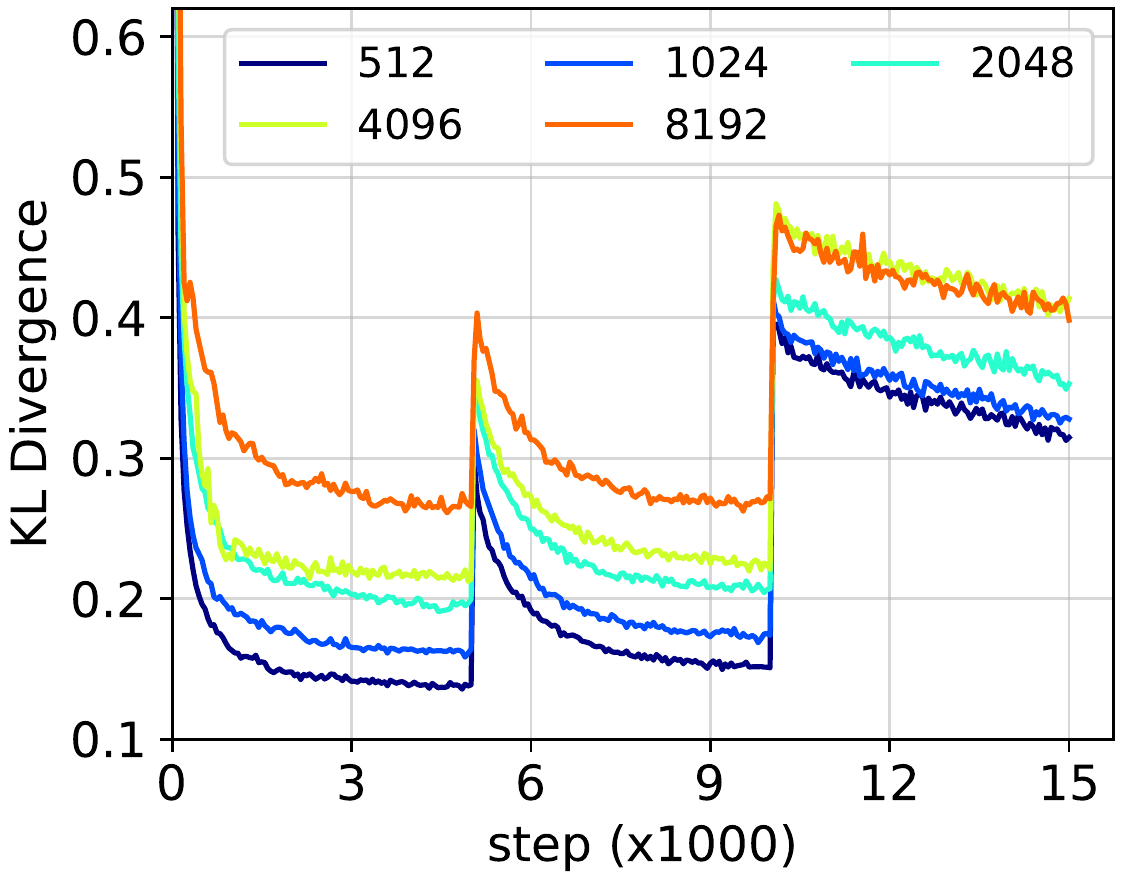}
\par\end{centering}
}\hspace{0.005\textwidth}\subfloat[$\protect\Loss_{\text{KD}}$ on samples from $\protect\Gen$\label{fig:KDLoss_on_gen}]{\begin{centering}
\includegraphics[width=0.3\textwidth]{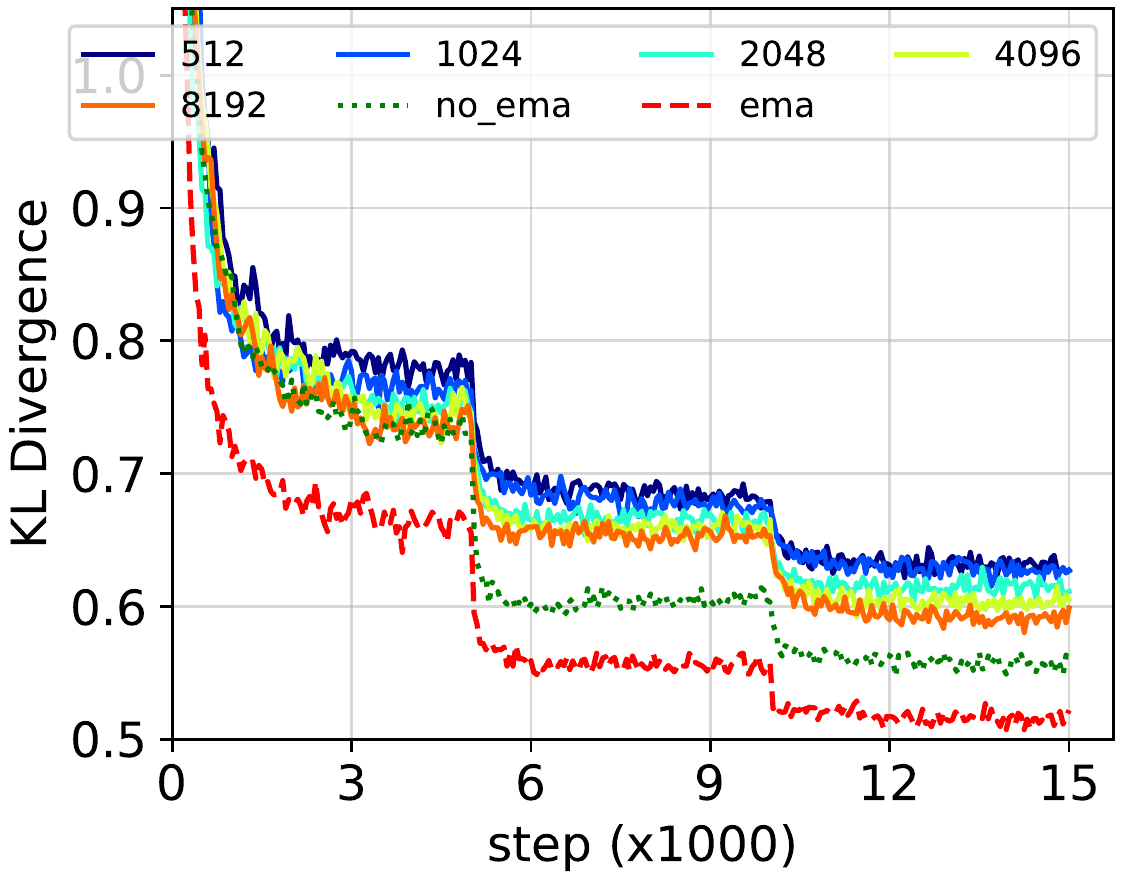}
\par\end{centering}
}\hspace{0.005\textwidth}\subfloat[$\protect\Loss_{\text{KD}}$ on samples from $\tilde{\protect\Gen}$\label{fig:KDLoss_on_ema}]{\begin{centering}
\includegraphics[width=0.3\textwidth]{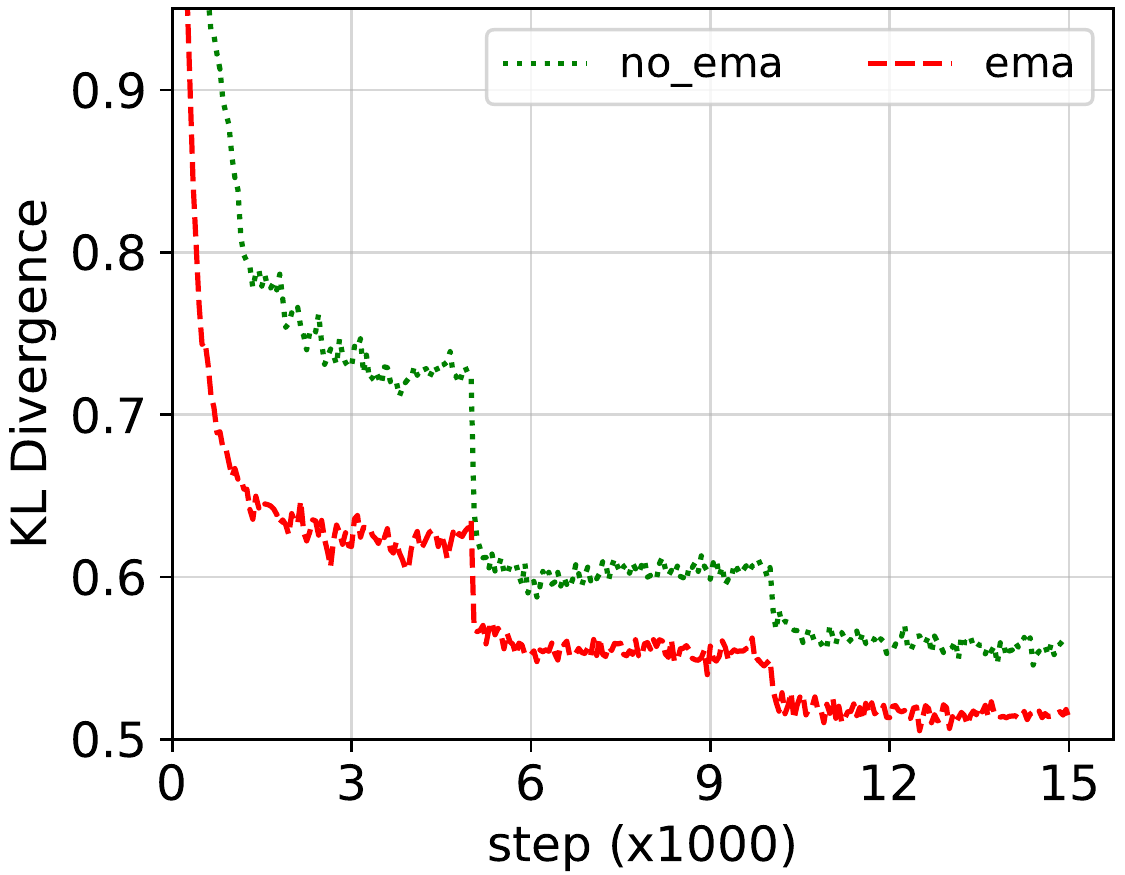}
\par\end{centering}
}
\par\end{centering}
\caption{Distillation loss curves on memory  samples (a) and samples generated
by $\protect\Gen$ (b) and $\tilde{\protect\Gen}$ (c). The numbers
in the legends denote DFKD-Mem with the corresponding memory sizes.
``no\_ema'' and ``ema'' denote ABM and $\protect\Model$, respectively.}
\end{figure}

In this experiment,  we found that DFKD-Mem often performs worse
than ABM on CIFAR100 and TinyImageNet. We found the decay of the student
learning rate is the main reason for this. As shown in Fig.~\ref{fig:KDLoss_on_que},
the distillation loss on memory samples surges when the (student)
learning rate is decayed and cannot recover if the new learning rate
is too small ($lr_{\Stu}$ = 1e-4), which is in contrast to the distillation
loss on synthetic samples from $\Gen$ or $\tilde{\Gen}$ (Figs.~\ref{fig:KDLoss_on_gen},\ref{fig:KDLoss_on_ema}).
This implies a potential issue of storing old samples in a memory
bank instead of using the EMA generator as memory samples could be
completely out-of-date if the student suddenly change its state (e.g.,
via learning rate decay). However, even when the learning rate does
not change (e.g., from step 0 to step 100 on CIFAR100 or on ImageNet/Places365),
DFKD-Mem still performs worse than our method.

\begin{table}
\begin{centering}
\begin{tabular}{c|cccccc}
Dataset & CIFAR10  & CIFAR100 & TinyIN & ImageNet & Places365 & Food101\tabularnewline
Arch. & $\diamondsuit$ & $\diamondsuit$ & $\heartsuit$ & $\clubsuit$ & $\clubsuit$ & $\clubsuit$\tabularnewline
\hline 
\hline 
Teacher & 94.65 & 75.65 & 66.47 & 56.52 & 50.80 & 65.15\tabularnewline
\hline 
ABM & 92.38 & 62.59 & 59.75 & 41.23 & 41.84 & 60.37\tabularnewline
DFKD-Mem & 92.09 & 61.25 & 58.66 & 43.30 & 42.38 & 61.25\tabularnewline
\hline 
$\Model$ & \textbf{92.64} & \textbf{64.05} & \textbf{62.32} & \textbf{45.48} & \textbf{43.67} & \textbf{61.74}\tabularnewline
\end{tabular}
\par\end{centering}
\caption{Classification accuracy (in \%) of $\protect\Model$ and its related
baselines on 3 small and 3 large image datasets. The teacher/student
architecture settings are ResNet34/ResNet18 ($\heartsuit$), WRN40-2/WRN16-2
($\diamondsuit$), and AlexNet/AlexNet ($\clubsuit$). The teacher's
results are from our own runs (Appdx.~\ref{subsec:Results-of-teacher}).
The best results are highlighted in \textbf{bold}.\label{tab:Results-Related-Baselines}}
\end{table}

\subsubsection{Comparing the changes in update of $\tilde{\protect\Gen}$ and $\protect\Gen$}

\begin{figure}
\begin{centering}
\subfloat[CIFAR100]{\begin{centering}
\includegraphics[width=0.22\textwidth]{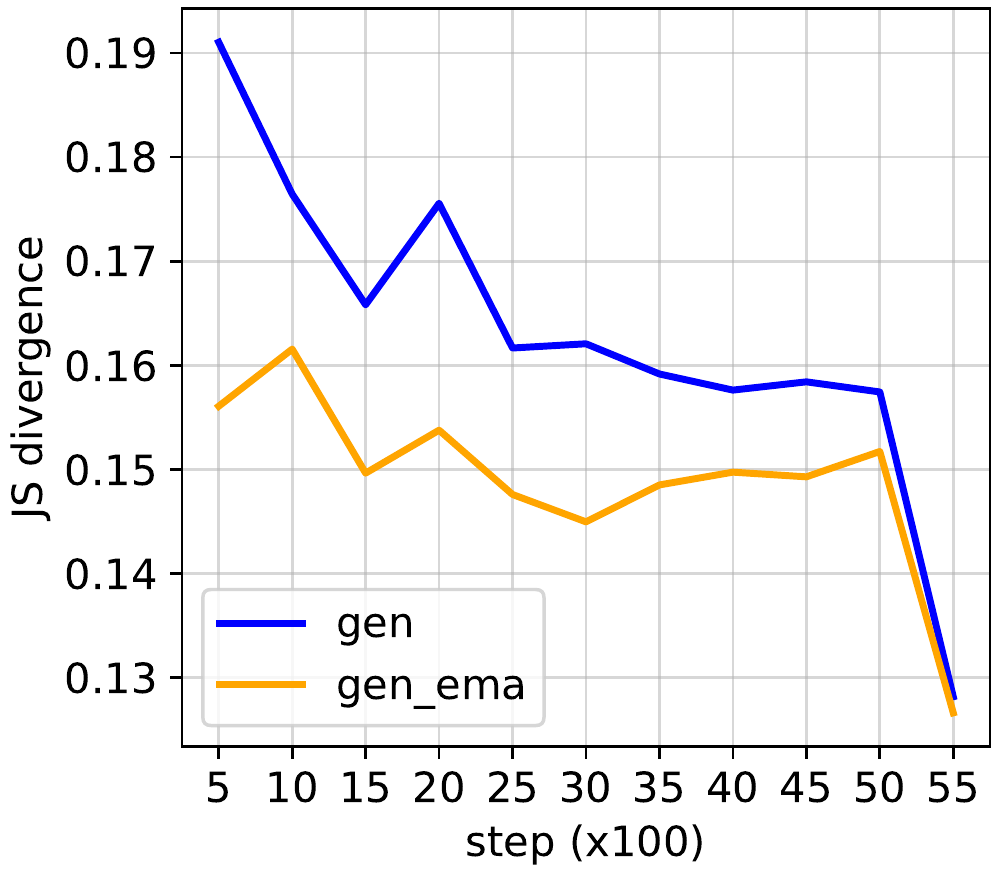}
\par\end{centering}
}\hspace{0.005\textwidth}\subfloat[TinyImageNet]{\begin{centering}
\includegraphics[width=0.22\textwidth]{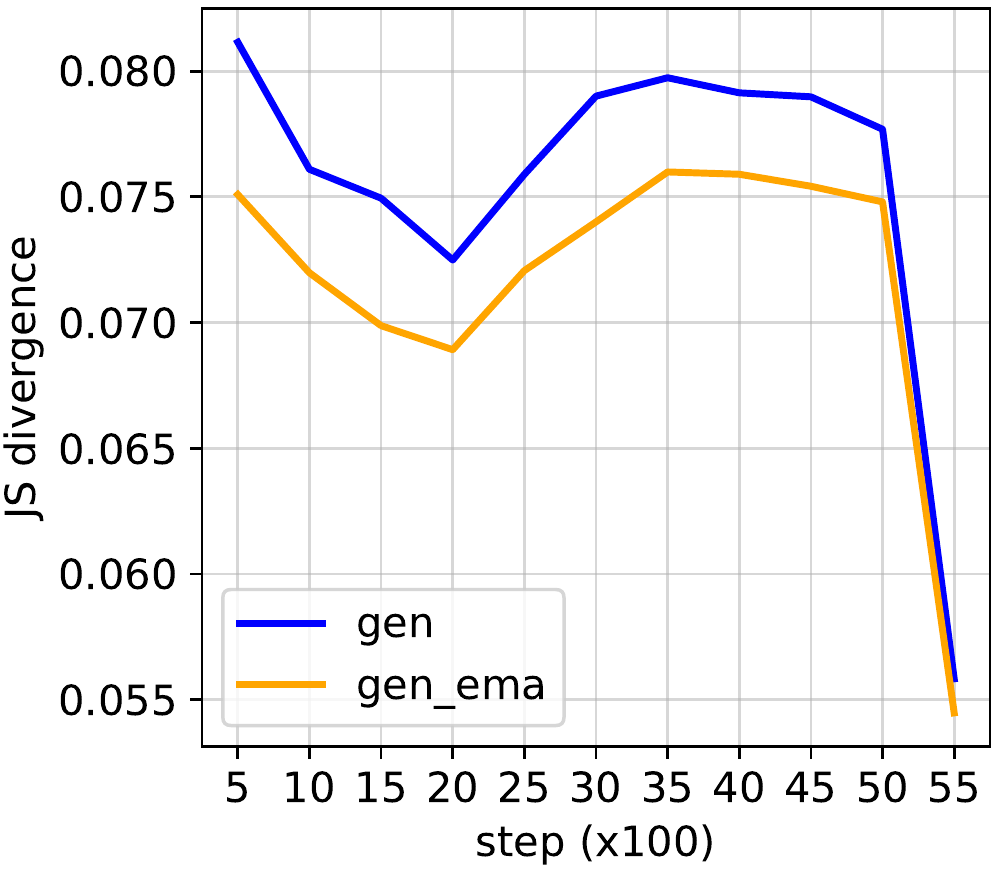}
\par\end{centering}
}\hspace{0.005\textwidth}\subfloat[ImageNet]{\begin{centering}
\includegraphics[width=0.22\textwidth]{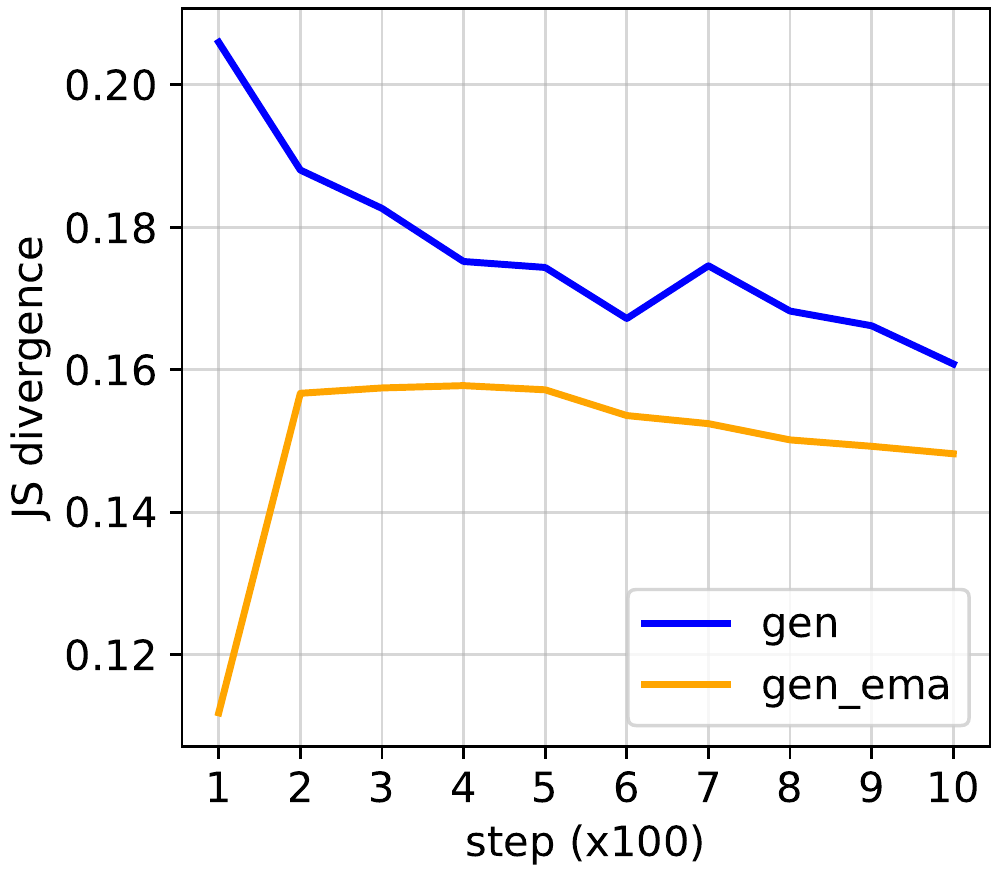}
\par\end{centering}
}\hspace{0.005\textwidth}\subfloat[Places365]{\begin{centering}
\includegraphics[width=0.22\textwidth]{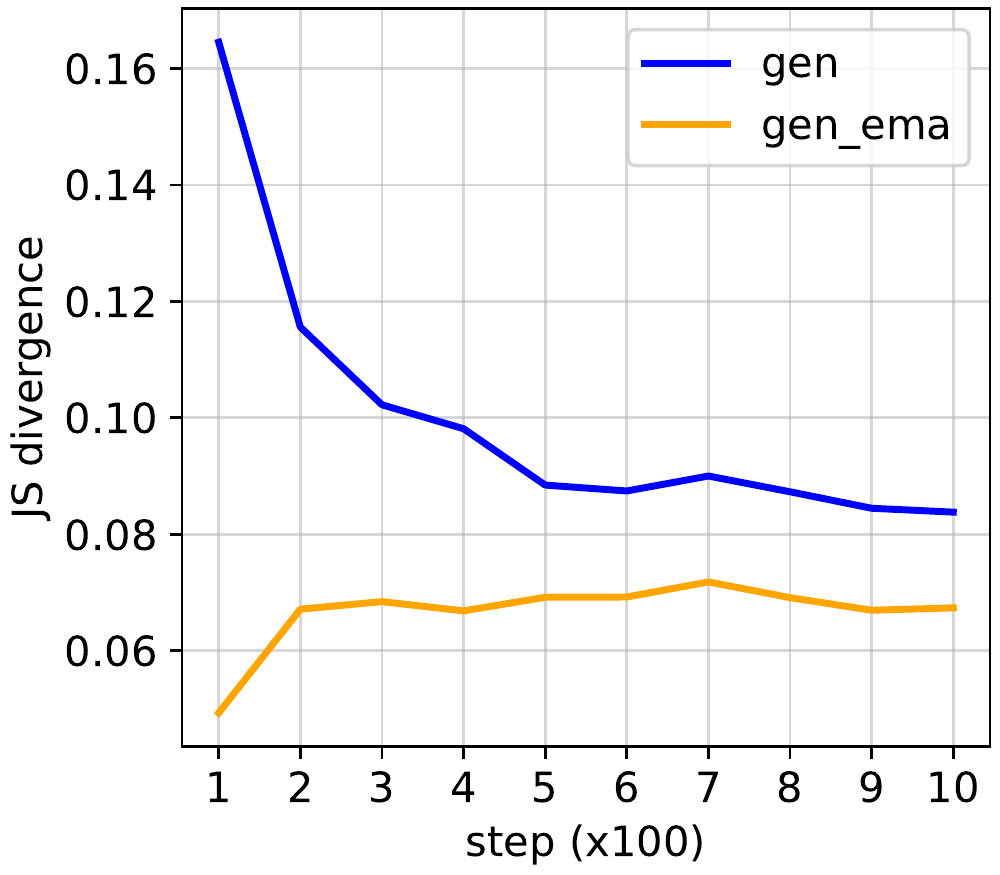}
\par\end{centering}
}
\par\end{centering}
\caption{Average Jensen-Shannon divergences between the prediction probabilities
of $\protect\Stu_{t-\tau}$ and $\protect\Tea$ computed on synthetic
samples from $\protect\Gen_{t}$ (labeled as ``gen'') and $\tilde{\protect\Gen}_{t}$
(labeled as ``gen\_ema'') for different datasets. For CIFAR100 and
TinyImageNet, we set $t\in[500,5500]$ with step size of 500 and $\tau=50$.
For ImageNet and Places365, we set $t\in[100,1000]$ with step size
of 100 and $\tau=10$. Note that the sudden drops at step 5000 in
(a), (b) correspond to the decay of the learning rate by 0.1.\label{fig:JSDiv_4_datasets}}
\end{figure}

In order to see whether $\tilde{\Gen}$ actually has smaller changes
in update than $\Gen$ or not, we perform the following experiment:
Let $\Gen_{t}$ and $\tilde{\Gen}_{t}$ be the versions of the generator
and the EMA generator respectively at step $t$, and $\Stu_{t-\tau}$
be the version of the student at step $t-\tau$ ($0<\tau<t$). We
then measure two different average Jensen-Shannon (JS) divergences
between the prediction probabilities of $\Stu_{t-\tau}$ and $\Tea$
on two separate sets of synthetic samples from $\Gen_{t}$ and $\tilde{\Gen}_{t}$.
We hypothesize that if $\Gen$ has a smaller change in the distribution
of synthetic samples than $\Gen$, $\Stu$ will memorize the samples
from $\tilde{\Gen}$ more and will match $\Tea$ better on those samples,
which leads to a smaller average JS divergence. This hypothesis is
clearly reflected on results in Fig.~\ref{fig:JSDiv_4_datasets},
which verifies the reasonability of using $\tilde{\Gen}$ in alleviating
the large distribution shift problem caused by $\Gen$.

\subsubsection{Visualization of synthetic samples\label{subsec:Visualization-of-synthetic-samples}}

\begin{figure}
\begin{centering}
\subfloat[CIFAR100]{\begin{centering}
\begin{tabular}{cc}
\includegraphics[width=0.2\textwidth]{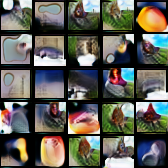} & \includegraphics[width=0.2\textwidth]{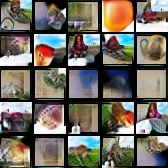}\tabularnewline
Generator & EMA Generator\tabularnewline
\end{tabular}
\par\end{centering}
}\hspace{0.01\textwidth}\subfloat[ImageNet]{\begin{centering}
\begin{tabular}{cc}
\includegraphics[width=0.2\textwidth]{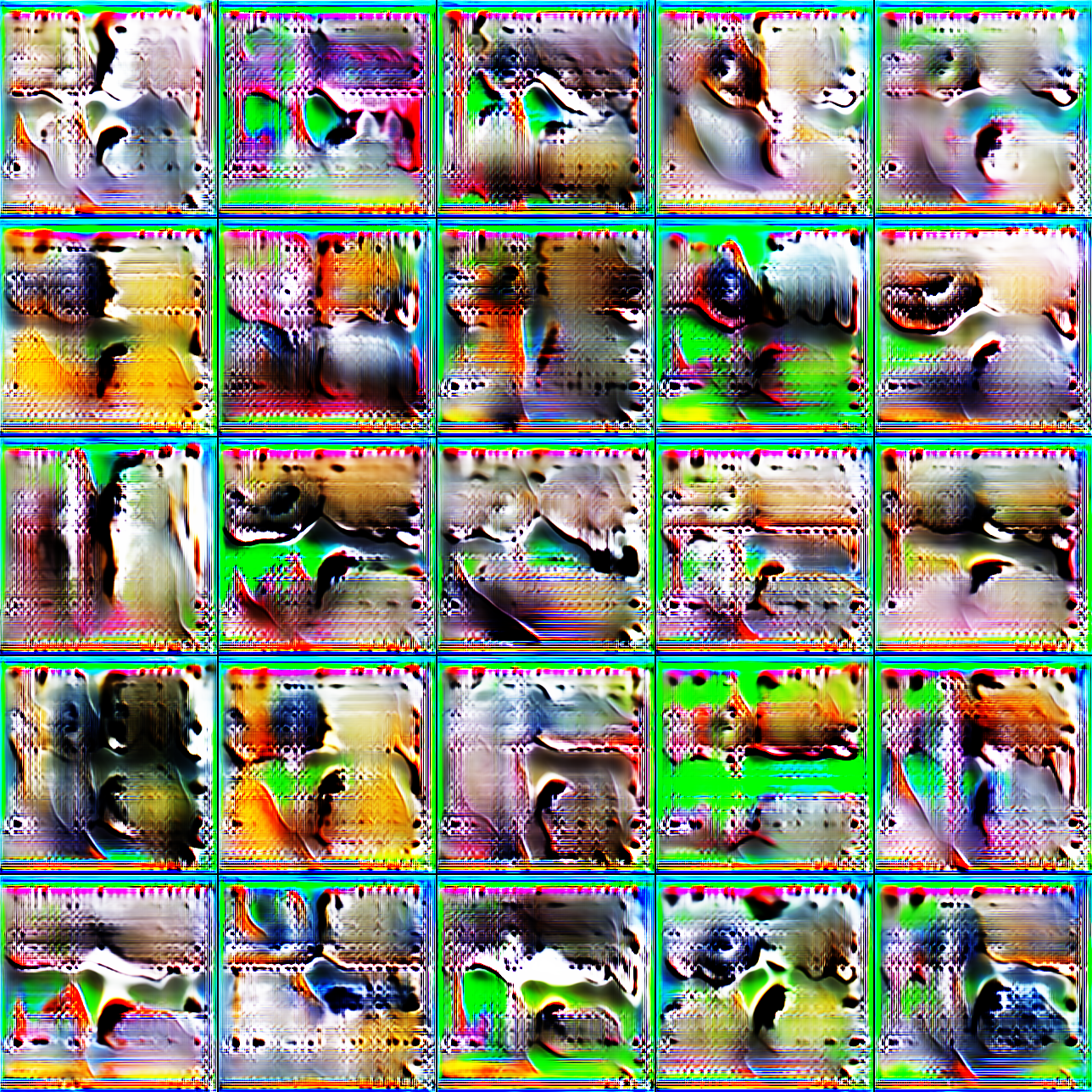} & \includegraphics[width=0.2\textwidth]{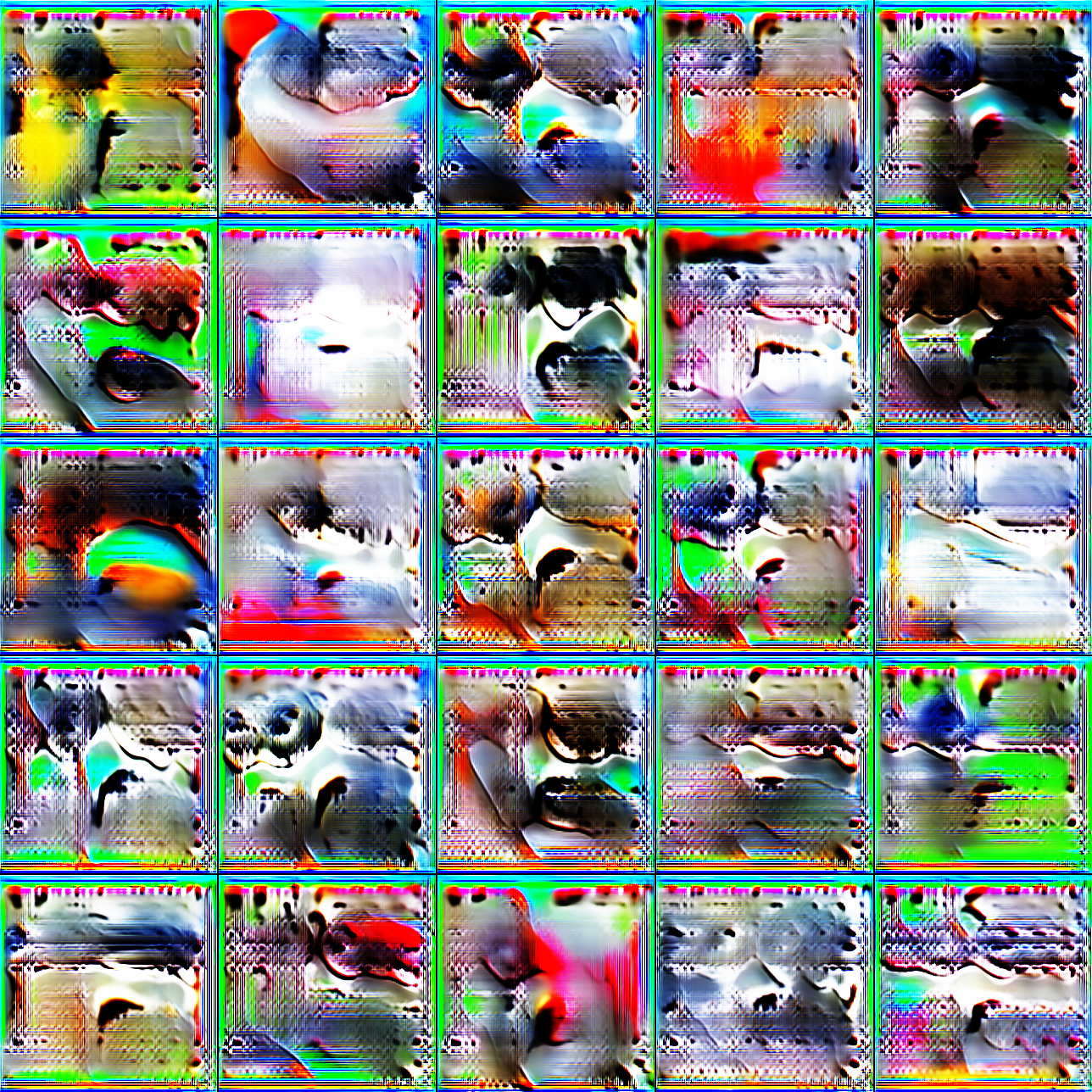}\tabularnewline
Generator & EMA Generator\tabularnewline
\end{tabular}
\par\end{centering}
}
\par\end{centering}
\caption{Synthetic data generated by $\protect\Gen$ and $\tilde{\protect\Gen}$
in case the original dataset is CIFAR100 (a) and ImageNet (b).\label{fig:syn-data}}
\end{figure}

In Fig.~\ref{fig:syn-data}, we visualize the synthetic data generated
by $\Gen$ and $\tilde{\Gen}$. Although the generated images are
not visually realistic, they are visually diverse, suggesting no mode
collapse has occurred during training $\Model$. Besides, samples
generated by $\tilde{\Gen}$ are different from those generated by
$\Gen$ which indicates that the EMA generator could act as a complement
for the generator in our model. 

\subsection{Sensitivity Analysis\label{subsec:Ablation-Study}}

Below we investigate some choices of hyperparameters that could affect
the performance of $\Model$. Unless stated otherwise, the dataset
we use is CIFAR100.

\paragraph{Different values of the coefficients in $\protect\Loss_{\text{\ensuremath{\protect\Stu}}}$}

\begin{figure}
\begin{centering}
\begin{minipage}[b]{0.24\textwidth}%
\begin{center}
\includegraphics[width=1\textwidth]{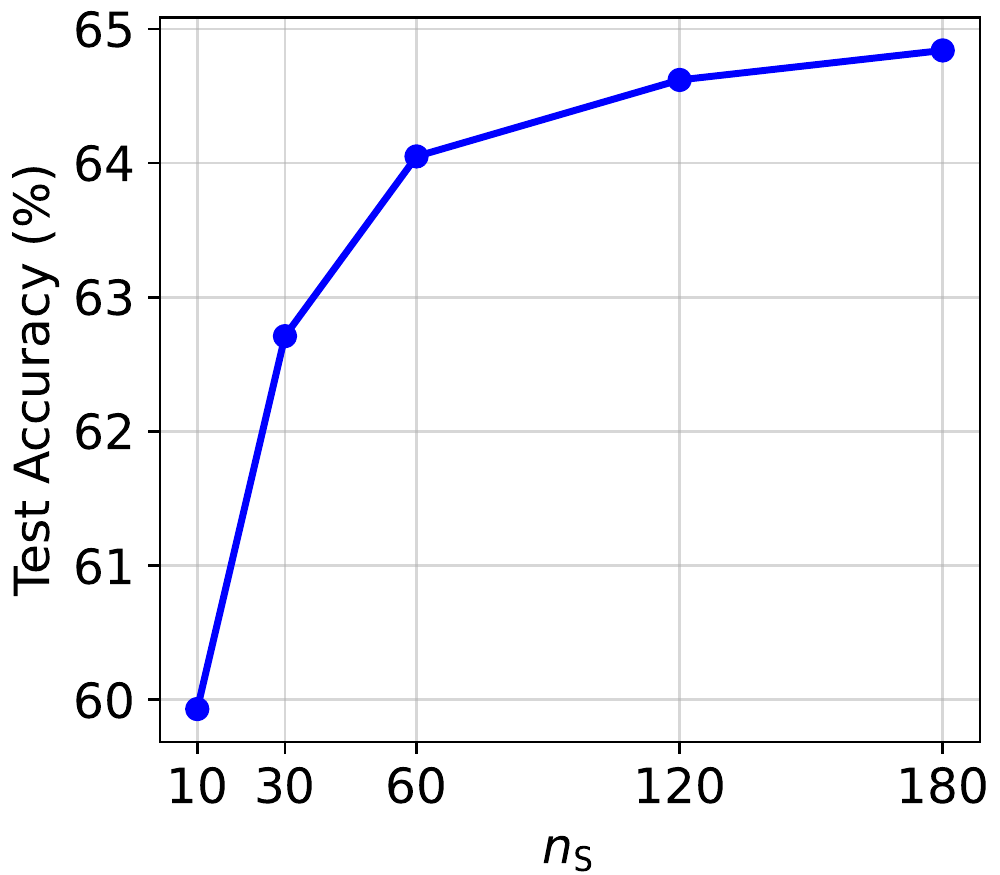}
\par\end{center}%
\end{minipage}\hspace{0.02\textwidth}%
\begin{minipage}[b]{0.24\textwidth}%
\begin{center}
\includegraphics[width=1\textwidth]{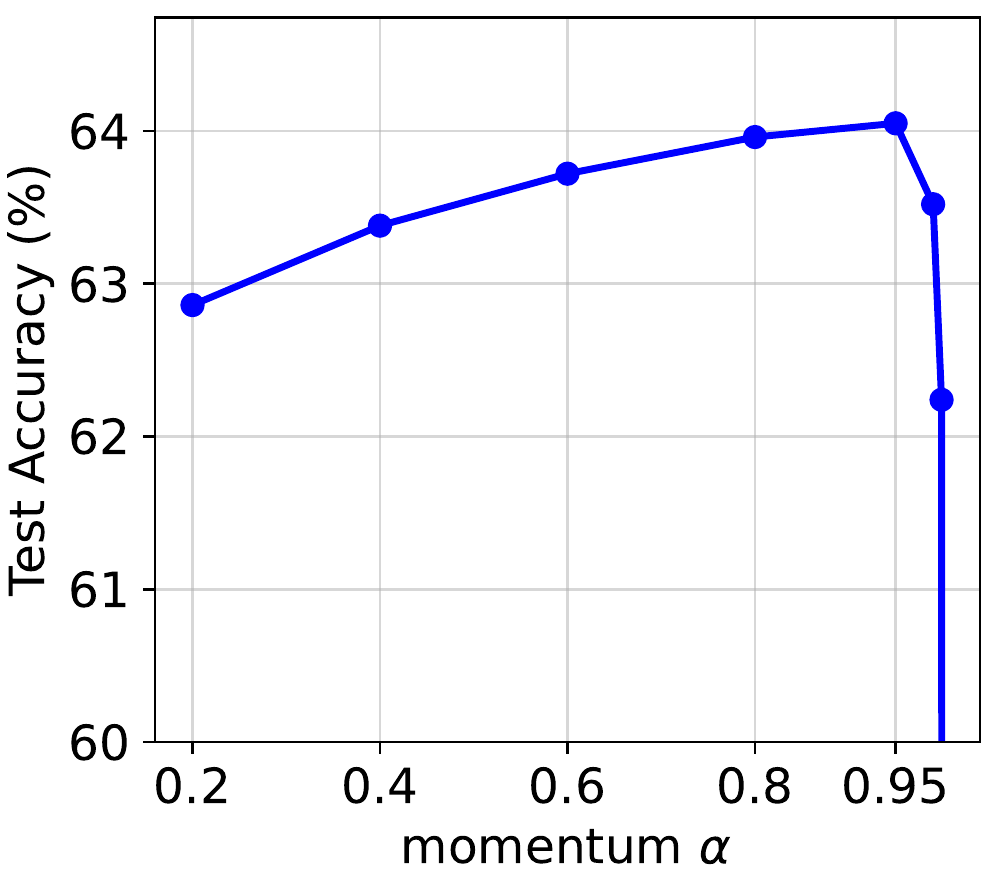}
\par\end{center}%
\end{minipage}\hspace{0.02\textwidth}%
\begin{minipage}[b]{0.4\textwidth}%
\begin{center}
\resizebox{\textwidth}{!}{
\begin{tabular}[b]{cc|cccc}
 & \multicolumn{1}{c}{} & \multicolumn{4}{c}{$\lambda_{0}$}\tabularnewline
 & \multicolumn{1}{c}{} & 1.0 & 0.3 & 0.1 & 0.0\tabularnewline
\cline{3-6} 
\multirow{4}{*}{$\lambda_{1}$} & 1.0 & \textbf{64.05} & 63.26 & 62.43 & 1.08\tabularnewline
 & 0.3 & 63.65 & - & - & -\tabularnewline
 & 0.1 & 62.80 & - & - & -\tabularnewline
 & 0.0 & 62.59 & - & - & -\tabularnewline
\end{tabular}}
\par\end{center}%
\end{minipage}
\par\end{centering}
\caption{Test accuracy of our method w.r.t. different values of student update
steps $n_{\protect\Stu}$ (\textbf{left}), of the momentum $\alpha$
(\textbf{middle}), and of the coefficients ($\lambda_{0}$, $\lambda_{1}$)
in Eq.~\ref{eq:MAD_LossS} (\textbf{right}). The dataset is CIFAR100.\label{fig:Abl-results}}
\end{figure}

We can control the relative importance of samples from $\tilde{\Gen}$
over those from $\Gen$ by changing the values of the two coefficients
$\lambda_{0}$ and $\lambda_{1}$ in $\Loss_{\Stu}$ (Eq.~\ref{eq:MAD_LossS}).
As shown in Fig.~\ref{fig:Abl-results} (right), the best result
is obtained when $\lambda_{0}=\lambda_{1}=1$ which is our default
setting for $\Model$. Decreasing either $\lambda_{0}$ or $\lambda_{1}$
will lead to worse performance. In the worst case when $\lambda_{0}=0$
and $\lambda_{1}=1$, the model learns for some epochs and then suddenly
stops learning. The main reason is that $\tilde{\Gen}$ often updates
much slower than $\Gen$ and $\Stu$ so we need adversarial samples
from $\Gen$ to keep $\Stu$ learning. Otherwise, $\Stu$ will overfit
the samples from $\tilde{\Gen}$ and learn nothing.

\paragraph{Different values of the momentum $\alpha$}

In Fig.~\ref{fig:Abl-results} (middle), we show the classification
accuracy of the student of $\Model$ with $\alpha$ in \{0.2, 0.4,
0.6, 0.8, 0.95, 0.99, 0.999, 1.0\}. We observe that the performance
of our method increases when $\alpha$ becomes larger as $\tilde{\Gen}$
is more different from $\Gen$. However, if $\alpha$ is too large
(e.g., 0.999), the performance drops since the update of $\Gen$ is
very small, resulting in almost no change in the distribution of synthetic
samples. In case $\alpha=1.0$, $\Model$ only achieves about 50\%
prediction accuracy on test data.

\paragraph{Different update steps of $\protect\Stu$}

From Fig.~\ref{fig:Abl-results} (left), we see that increasing the
number of update steps for $\Stu$ usually leads to better performances
of $\Model$ since $\Stu$ learns to match $\Tea$ better. In exchange,
the training time will increase.

\section{Conclusion\label{sec:Conclusion}}

We have presented Momentum Adversarial Distillation ($\Model$), a
simple yet effective method to deal with the large distribution shift
problem in adversarial Data-Free Knowledge Distillation (DFKD). $\Model$
maintains an exponential moving average (EMA) copy of the generator
which, by design, encapsulates information about past updates of the
generator and is updated at a slower pace than the generator. By training
the student on samples from both the generator and the EMA generator,
$\Model$ can prevent the student from adapting too much to the generator
at the current step and forgetting old knowledge learned from the
generator at the previous steps. We have also described a new type
of conditional generator along with a new loss for training it which
enable our model to learn well on large datasets. Our experiments
on various datasets demonstrated the superior performance of our method
over competing baselines that either use only a generator or use a
memory bank in place of an EMA generator.

We note that our idea of using an EMA generator to mitigate large
distribution shifts is general and can be generalized to other machine
learning problems besides DFKD. For example, the technique can be
adapted for general continual learning (in which large distribution
shifts can happen gradually or suddenly with no clear task boundaries),
and source-data-free domain adaptation. Our method is also well suited
for DFKD with other data types such as video, text, or graph.

\paragraph{Limitations}

In our current implementation of $\Model$, we perform an additional
forward pass through $\tilde{\Gen}$ for every training step of $\Stu$.
This increases the total training time of by about 40\% compared to
ABM. However, this technical problem can be somewhat addressed by
first storing synthetic samples from $\Gen$ and $\tilde{\Gen}$ in
a buffer before each training stage of $\Stu$ and then training $\Stu$
with samples from the buffer only. This will be left for future work.

\paragraph{Negative Social Impacts}

The DFKD problem may have negative social impacts related to data
privacy as generated data could somehow reveal the original training
data. However, as shown in Fig.~\ref{fig:syn-data}, our proposed
method does not attempt to improve the visual interpretability of
synthetic data but addresses the large distribution shift problem.
This target seems to be harmless to the society.

\section*{Acknowledgement}

This research was partially funded by the Australian Government through
the Australian Research Council (ARC). Prof. Venkatesh is the recipient
of an ARC Australian Laureate Fellowship (FL170100006).

\bibliographystyle{plain}
\bibliography{CAD}

\begin{thebibliography}{10}

\bibitem{ahn2019variational}
Sungsoo Ahn, Shell~Xu Hu, Andreas Damianou, Neil~D Lawrence, and Zhenwen Dai.
\newblock Variational information distillation for knowledge transfer.
\newblock In {\em Proceedings of the IEEE/CVF Conference on Computer Vision and
  Pattern Recognition}, pages 9163--9171, 2019.

\bibitem{binici2022robust}
Kuluhan Binici, Shivam Aggarwal, Nam~Trung Pham, Karianto Leman, and Tulika
  Mitra.
\newblock Robust and resource-efficient data-free knowledge distillation by
  generative pseudo replay.
\newblock {\em arXiv preprint arXiv:2201.03019}, 2022.

\bibitem{binici2022preventing}
Kuluhan Binici, Nam~Trung Pham, Tulika Mitra, and Karianto Leman.
\newblock Preventing catastrophic forgetting and distribution mismatch in
  knowledge distillation via synthetic data.
\newblock In {\em Proceedings of the IEEE/CVF Winter Conference on Applications
  of Computer Vision}, pages 663--671, 2022.

\bibitem{bossard14}
Lukas Bossard, Matthieu Guillaumin, and Luc Van~Gool.
\newblock Food-101 - mining discriminative components with random forests.
\newblock In {\em European Conference on Computer Vision}, 2014.

\bibitem{carion2020end}
Nicolas Carion, Francisco Massa, Gabriel Synnaeve, Nicolas Usunier, Alexander
  Kirillov, and Sergey Zagoruyko.
\newblock End-to-end object detection with transformers.
\newblock In {\em European conference on computer vision}, pages 213--229.
  Springer, 2020.

\bibitem{chawla2021data}
Akshay Chawla, Hongxu Yin, Pavlo Molchanov, and Jose Alvarez.
\newblock Data-free knowledge distillation for object detection.
\newblock In {\em Proceedings of the IEEE/CVF Winter Conference on Applications
  of Computer Vision}, pages 3289--3298, 2021.

\bibitem{chen2019data}
Hanting Chen, Yunhe Wang, Chang Xu, Zhaohui Yang, Chuanjian Liu, Boxin Shi,
  Chunjing Xu, Chao Xu, and Qi~Tian.
\newblock Data-free learning of student networks.
\newblock In {\em Proceedings of the IEEE/CVF International Conference on
  Computer Vision}, pages 3514--3522, 2019.

\bibitem{choi2020data}
Yoojin Choi, Jihwan Choi, Mostafa El-Khamy, and Jungwon Lee.
\newblock Data-free network quantization with adversarial knowledge
  distillation.
\newblock In {\em Proceedings of the IEEE/CVF Conference on Computer Vision and
  Pattern Recognition Workshops}, pages 710--711, 2020.

\bibitem{deng2009imagenet}
Jia Deng, Wei Dong, Richard Socher, Li-Jia Li, Kai Li, and Li~Fei-Fei.
\newblock Imagenet: A large-scale hierarchical image database.
\newblock In {\em 2009 IEEE Conference on Computer Vision and Pattern
  Recognition}, pages 248--255. IEEE, 2009.

\bibitem{devlin2018bert}
Jacob Devlin, Ming-Wei Chang, Kenton Lee, and Kristina Toutanova.
\newblock Bert: Pre-training of deep bidirectional transformers for language
  understanding.
\newblock {\em arXiv preprint arXiv:1810.04805}, 2018.

\bibitem{fang2021mosaicking}
Gongfan Fang, Yifan Bao, Jie Song, Xinchao Wang, Donglin Xie, Chengchao Shen,
  and Mingli Song.
\newblock Mosaicking to distill: Knowledge distillation from out-of-domain
  data.
\newblock {\em Advances in Neural Information Processing Systems},
  34:11920--11932, 2021.

\bibitem{fang2021up}
Gongfan Fang, Kanya Mo, Xinchao Wang, Jie Song, Shitao Bei, Haofei Zhang, and
  Mingli Song.
\newblock Up to 100x faster data-free knowledge distillation.
\newblock {\em arXiv preprint arXiv:2112.06253}, 2021.

\bibitem{fang2019data}
Gongfan Fang, Jie Song, Chengchao Shen, Xinchao Wang, Da~Chen, and Mingli Song.
\newblock Data-free adversarial distillation.
\newblock {\em arXiv preprint arXiv:1912.11006}, 2019.

\bibitem{fang2021contrastive}
Gongfan Fang, Jie Song, Xinchao Wang, Chengchao Shen, Xingen Wang, and Mingli
  Song.
\newblock Contrastive model inversion for data-free knowledge distillation.
\newblock {\em arXiv preprint arXiv:2105.08584}, 2021.

\bibitem{goyal2017accurate}
Priya Goyal, Piotr Doll{\'a}r, Ross Girshick, Pieter Noordhuis, Lukasz
  Wesolowski, Aapo Kyrola, Andrew Tulloch, Yangqing Jia, and Kaiming He.
\newblock Accurate, large minibatch sgd: Training imagenet in 1 hour.
\newblock {\em arXiv preprint arXiv:1706.02677}, 2017.

\bibitem{gulrajani2017improved}
Ishaan Gulrajani, Faruk Ahmed, Martin Arjovsky, Vincent Dumoulin, and Aaron~C
  Courville.
\newblock Improved training of wasserstein gans.
\newblock {\em Advances in neural information processing systems}, 30, 2017.

\bibitem{han2021robustness}
Pengchao Han, Jihong Park, Shiqiang Wang, and Yejun Liu.
\newblock Robustness and diversity seeking data-free knowledge distillation.
\newblock In {\em ICASSP 2021-2021 IEEE International Conference on Acoustics,
  Speech and Signal Processing (ICASSP)}, pages 2740--2744. IEEE, 2021.

\bibitem{he2016deep}
Kaiming He, Xiangyu Zhang, Shaoqing Ren, and Jian Sun.
\newblock Deep residual learning for image recognition.
\newblock In {\em Proceedings of the IEEE conference on computer vision and
  pattern recognition}, pages 770--778, 2016.

\bibitem{hinton2015distilling}
Geoffrey Hinton, Oriol Vinyals, and Jeff Dean.
\newblock Distilling the knowledge in a neural network.
\newblock {\em arXiv preprint arXiv:1503.02531}, 2015.

\bibitem{ioffe2015batch}
Sergey Ioffe and Christian Szegedy.
\newblock Batch normalization: Accelerating deep network training by reducing
  internal covariate shift.
\newblock In {\em International conference on machine learning}, pages
  448--456. PMLR, 2015.

\bibitem{Kingma_Ba_14Adam}
Diederik~P Kingma and Jimmy Ba.
\newblock Adam: A {M}ethod for {S}tochastic {O}ptimization.
\newblock {\em arXiv preprint arXiv:1412.6980}, 2014.

\bibitem{kingma2013auto}
Diederik~P Kingma and Max Welling.
\newblock Auto-encoding variational bayes.
\newblock {\em arXiv preprint arXiv:1312.6114}, 2013.

\bibitem{kirkpatrick2017overcoming}
James Kirkpatrick, Razvan Pascanu, Neil Rabinowitz, Joel Veness, Guillaume
  Desjardins, Andrei~A Rusu, Kieran Milan, John Quan, Tiago Ramalho, Agnieszka
  Grabska-Barwinska, et~al.
\newblock Overcoming catastrophic forgetting in neural networks.
\newblock {\em Proceedings of the national academy of sciences},
  114(13):3521--3526, 2017.

\bibitem{krizhevsky2009learning}
Alex Krizhevsky.
\newblock Learning multiple layers of features from tiny images.
\newblock Technical report, 2009.

\bibitem{le2015tiny}
Ya~Le and Xuan Yang.
\newblock Tiny imagenet visual recognition challenge.
\newblock Technical report, 2015.

\bibitem{liang2018generative}
Kevin~J Liang, Chunyuan Li, Guoyin Wang, and Lawrence Carin.
\newblock Generative adversarial network training is a continual learning
  problem.
\newblock {\em arXiv preprint arXiv:1811.11083}, 2018.

\bibitem{lopes2017data}
Raphael~Gontijo Lopes, Stefano Fenu, and Thad Starner.
\newblock Data-free knowledge distillation for deep neural networks.
\newblock {\em arXiv preprint arXiv:1710.07535}, 2017.

\bibitem{lucas2019understanding}
James Lucas, George Tucker, Roger Grosse, and Mohammad Norouzi.
\newblock Understanding posterior collapse in generative latent variable
  models.
\newblock 2019.

\bibitem{luo2020large}
Liangchen Luo, Mark Sandler, Zi~Lin, Andrey Zhmoginov, and Andrew Howard.
\newblock Large-scale generative data-free distillation.
\newblock {\em arXiv preprint arXiv:2012.05578}, 2020.

\bibitem{mao2019mode}
Qi~Mao, Hsin-Ying Lee, Hung-Yu Tseng, Siwei Ma, and Ming-Hsuan Yang.
\newblock Mode seeking generative adversarial networks for diverse image
  synthesis.
\newblock In {\em Proceedings of the IEEE/CVF conference on computer vision and
  pattern recognition}, pages 1429--1437, 2019.

\bibitem{micaelli2019zero}
Paul Micaelli and Amos~J Storkey.
\newblock Zero-shot knowledge transfer via adversarial belief matching.
\newblock {\em Advances in Neural Information Processing Systems}, 32, 2019.

\bibitem{mordvintsev2015inceptionism}
Alexander Mordvintsev, Christopher Olah, and Mike Tyka.
\newblock Inceptionism: Going deeper into neural networks.
\newblock 2015.

\bibitem{nayak2021effectiveness}
Gaurav~Kumar Nayak, Konda~Reddy Mopuri, and Anirban Chakraborty.
\newblock Effectiveness of arbitrary transfer sets for data-free knowledge
  distillation.
\newblock In {\em Proceedings of the IEEE/CVF Winter Conference on Applications
  of Computer Vision}, pages 1430--1438, 2021.

\bibitem{nayak2019zero}
Gaurav~Kumar Nayak, Konda~Reddy Mopuri, Vaisakh Shaj, Venkatesh~Babu
  Radhakrishnan, and Anirban Chakraborty.
\newblock Zero-shot knowledge distillation in deep networks.
\newblock In {\em International Conference on Machine Learning}, pages
  4743--4751. PMLR, 2019.

\bibitem{park2019relational}
Wonpyo Park, Dongju Kim, Yan Lu, and Minsu Cho.
\newblock Relational knowledge distillation.
\newblock In {\em Proceedings of the IEEE/CVF Conference on Computer Vision and
  Pattern Recognition}, pages 3967--3976, 2019.

\bibitem{passalis2018learning}
Nikolaos Passalis and Anastasios Tefas.
\newblock Learning deep representations with probabilistic knowledge transfer.
\newblock In {\em Proceedings of the European Conference on Computer Vision
  (ECCV)}, pages 268--284, 2018.

\bibitem{qu2021enhancing}
Xiaoyang Qu, Jianzong Wang, and Jing Xiao.
\newblock Enhancing data-free adversarial distillation with activation
  regularization and virtual interpolation.
\newblock In {\em ICASSP 2021-2021 IEEE International Conference on Acoustics,
  Speech and Signal Processing (ICASSP)}, pages 3340--3344. IEEE, 2021.

\bibitem{radford2021learning}
Alec Radford, Jong~Wook Kim, Chris Hallacy, Aditya Ramesh, Gabriel Goh,
  Sandhini Agarwal, Girish Sastry, Amanda Askell, Pamela Mishkin, Jack Clark,
  et~al.
\newblock Learning transferable visual models from natural language
  supervision.
\newblock In {\em International Conference on Machine Learning}, pages
  8748--8763. PMLR, 2021.

\bibitem{smith2021always}
James Smith, Yen-Chang Hsu, Jonathan Balloch, Yilin Shen, Hongxia Jin, and
  Zsolt Kira.
\newblock Always be dreaming: A new approach for data-free class-incremental
  learning.
\newblock In {\em Proceedings of the IEEE/CVF International Conference on
  Computer Vision}, pages 9374--9384, 2021.

\bibitem{thanh2020catastrophic}
Hoang Thanh-Tung and Truyen Tran.
\newblock Catastrophic forgetting and mode collapse in gans.
\newblock In {\em 2020 International Joint Conference on Neural Networks
  (IJCNN)}, pages 1--10. IEEE, 2020.

\bibitem{tian2019contrastive}
Yonglong Tian, Dilip Krishnan, and Phillip Isola.
\newblock Contrastive representation distillation.
\newblock In {\em International Conference on Learning Representations}, 2019.

\bibitem{tung2019similarity}
Frederick Tung and Greg Mori.
\newblock Similarity-preserving knowledge distillation.
\newblock In {\em Proceedings of the IEEE/CVF International Conference on
  Computer Vision}, pages 1365--1374, 2019.

\bibitem{vaswani2017attention}
Ashish Vaswani, Noam Shazeer, Niki Parmar, Jakob Uszkoreit, Llion Jones,
  Aidan~N Gomez, {\L}ukasz Kaiser, and Illia Polosukhin.
\newblock Attention is all you need.
\newblock {\em Advances in neural information processing systems}, 30, 2017.

\bibitem{wang2021data}
Zi~Wang.
\newblock Data-free knowledge distillation with soft targeted transfer set
  synthesis.
\newblock In {\em Proceedings of the AAAI Conference on Artificial
  Intelligence}, volume~35, pages 10245--10253, 2021.

\bibitem{yin2020dreaming}
Hongxu Yin, Pavlo Molchanov, Jose~M Alvarez, Zhizhong Li, Arun Mallya, Derek
  Hoiem, Niraj~K Jha, and Jan Kautz.
\newblock Dreaming to distill: Data-free knowledge transfer via deepinversion.
\newblock In {\em Proceedings of the IEEE/CVF Conference on Computer Vision and
  Pattern Recognition}, pages 8715--8724, 2020.

\bibitem{yoo2019knowledge}
Jaemin Yoo, Minyong Cho, Taebum Kim, and U~Kang.
\newblock Knowledge extraction with no observable data.
\newblock {\em Advances in Neural Information Processing Systems}, 32, 2019.

\bibitem{zagoruyko2016paying}
Sergey Zagoruyko and Nikos Komodakis.
\newblock Paying more attention to attention: Improving the performance of
  convolutional neural networks via attention transfer.
\newblock {\em arXiv preprint arXiv:1612.03928}, 2016.

\bibitem{zagoruyko2016wide}
Sergey Zagoruyko and Nikos Komodakis.
\newblock Wide residual networks.
\newblock In {\em British Machine Vision Conference 2016}. British Machine
  Vision Association, 2016.

\bibitem{zenke2017continual}
Friedemann Zenke, Ben Poole, and Surya Ganguli.
\newblock Continual learning through synaptic intelligence.
\newblock In {\em International Conference on Machine Learning}, pages
  3987--3995. PMLR, 2017.

\bibitem{zhao2021dual}
Haoran Zhao, Xin Sun, Junyu Dong, Milos Manic, Huiyu Zhou, and Hui Yu.
\newblock Dual discriminator adversarial distillation for data-free model
  compression.
\newblock {\em International Journal of Machine Learning and Cybernetics},
  pages 1--18, 2021.

\bibitem{zhou2017places}
Bolei Zhou, Agata Lapedriza, Aditya Khosla, Aude Oliva, and Antonio Torralba.
\newblock Places: A 10 million image database for scene recognition.
\newblock {\em IEEE Transactions on Pattern Analysis and Machine Intelligence},
  2017.

\end{thebibliography}

\newpage{}

\appendix

\section{Experimental Setup}

\subsection{Datasets\label{subsec:Datasets}}

In Table~\ref{tab:Dataset-Description}, we provide information about
the image size, the number of classes, and the number of training/test
samples of the datasets used in our experiment. 

\begin{table}[H]
\begin{centering}
\begin{tabular}{|c|c|c|c|c|}
\hline 
Datatset & Image size & \#classes & \#train & \#test\tabularnewline
\hline 
\hline 
CIFAR10 & \multirow{2}{*}{$3\times32\times32$} & 10 & 50,000 & 10,000\tabularnewline
\cline{1-1} \cline{3-5} 
CIFAR100 &  & 100 & 50,000  & 10,000\tabularnewline
\hline 
Tiny-ImageNet & $3\times64\times64$ & 200 & 100,000 & 10,000\tabularnewline
\hline 
ImageNet & \multirow{3}{*}{$3\times256\times256$} & 1000 & 1,281,167 & 50,000\tabularnewline
\cline{1-1} \cline{3-5} 
Places365 &  & 365 & 1,803,460 & 36,500\tabularnewline
\cline{1-1} \cline{3-5} 
Food101 &  & 101 & 75,750 & 25,250\tabularnewline
\hline 
\end{tabular}
\par\end{centering}
\caption{Details of the datasets used in our experiment.\label{tab:Dataset-Description}}
\end{table}

\subsection{Training Settings of Teacher\label{subsec:Teacher-Training-Settings}}

We provide training settings of the teacher w.r.t. different datasets
in Table~\ref{tab:Teacher-Settings}.

\begin{table}[H]
\begin{centering}
\begin{tabular}{|c|c|c|c|c|c|c|c|c|c|c|}
\hline 
\multirow{2}{*}{Dataset} & \multicolumn{10}{c|}{Training settings}\tabularnewline
\cline{2-11} 
 & $opt$ & $lr$ & $wd$ & $mo$ & $bs$ & $ls$ & $ld$ & $ldep$ & $ep$ & $wep$\tabularnewline
\hline 
\hline 
CIFAR10/100 & \multirow{4}{*}{SGD} & \multirow{2}{*}{0.1} & \multirow{2}{*}{5e-4} & \multirow{4}{*}{0.9} & 128 & No & \multirow{4}{*}{0.1} & \multirow{3}{*}{80, 120} & \multirow{3}{*}{160} & \multirow{2}{*}{0}\tabularnewline
\cline{1-1} \cline{6-7} 
TinyImageNet &  &  &  &  & 256 & No &  &  &  & \tabularnewline
\cline{1-1} \cline{3-4} \cline{6-7} \cline{11-11} 
Food101 &  & \multirow{2}{*}{0.01} & \multirow{2}{*}{1e-4} &  & 512 & Yes &  &  &  & \multirow{2}{*}{5}\tabularnewline
\cline{1-1} \cline{6-7} \cline{9-10} 
Places365 &  &  &  &  & 1024 & Yes &  & 30, 60 & 90 & \tabularnewline
\hline 
\end{tabular}
\par\end{centering}
\caption{Training settings of teacher w.r.t. different datasets. Meanings of
abbreviations: $opt$: optimizer, $lr$: learning rate, $wd$: weight
decay, $mo$: momentum, $bs$: batch size, $ls$: scaling learning
rate or not with the base batch size of 256 \cite{goyal2017accurate},
$ld$: learning rate decay, $ldep$: epochs at which learning rate
are decayed, $ep$: total number of epochs, $wep$: number of warm-up
epochs.\label{tab:Teacher-Settings}}
\end{table}

\subsection{Training Settings of $\protect\Model$\label{subsec:Training-Settings-of-Our-Method}}

In Tables~\ref{tab:Opt-Settings-Stu-Gen},\ref{tab:Train-Settings-Stu-Gen},\ref{tab:MAD_Coeffs},
we provide the training settings of $\Model$ used in this paper.
Despite multiple attempts, we could not find a global configuration
that works well for all datasets and architectures. 

In practice, we do \emph{not} optimize the student and the generator
via the plain losses in Eq.~\ref{eq:MAD_LossS} and Eq.~\ref{eq:ClassCondGen_loss},
respectively but with some additional regularizations on the output
logits of $\Tea$, $\Stu$ and $\Gen$. This prevents our losses from
being NaN when the logits grow too big. Specifically, we define $\Loss_{\Stu}$
as follows:
\begin{align}
\Loss_{\Stu}\triangleq\  & \lambda_{0}\Expect_{z\sim p(z),x=\Gen(z)}\left[\Loss_{\text{KD}}(x)+\zeta_{0}\max(\left|\Stu(x)\right|-\delta,0)\right]+\nonumber \\
 & \lambda_{1}\Expect_{z'\sim p(z),x'=\tilde{\Gen}(z')}\left[\Loss_{\text{KD}}(x')+\zeta_{0}\max(\left|\Stu(x')\right|-\delta,0)\right]\label{eq:LossStu_Practice}
\end{align}
where $\max(\left|\Stu(\cdot)\right|-\delta,0)$ ensures that the
output logit of $\Stu$ is between $[-\delta,\delta]$; $\zeta_{0}\ge0$
is a coefficient. 

And we define $\Loss_{\Gen,\Emb}$ as follows:
\begin{align}
\Loss_{\Gen,\Emb}\triangleq\  & \Expect_{z\sim\mathcal{N}(0,\mathrm{I}),y\sim\text{Cat}(C),u=\Gen_{\text{lg}}(z+y),x=\sigma(u)}\bigg[\nonumber \\
 & \ \ \ \ \ \ \ \ \ \ \ \ \ \ \ \ \ \ \ -\lambda_{2}\Loss_{\text{KD}}(x)+\lambda_{3}\Loss_{\text{NLL}}(x,y)+\lambda_{4}\Loss_{\text{NormReg}}(e_{y})\nonumber \\
 & \ \ \ \ \ \ \ \ \ \ \ \ \ \ \ \ \ \ \ +\zeta_{1}\max(|\Tea(x)|-\delta,0)+\zeta_{2}\max(|u|-\nu,0)\bigg]+\lambda_{5}\Loss_{\text{BNmm}}\label{eq:LossGen_Practice}
\end{align}
where $\Gen_{\text{lg}}$ denotes the generator that produces logits
instead of normalized images; $\sigma(\cdot)$ denotes the sigmoid
function; $\max(|u|-\nu,0)$ ensures that the output logit $u$ of
$\Gen_{\text{lg}}$ is between $[-\nu,\nu]$; $\max(|\Tea(x)|-\delta,0)$
ensures that the output logit of $\Tea$ w.r.t. the synthetic sample
$x$ is between $[-\delta,\delta]$; $\zeta_{1},\zeta_{2}\geq0$ are
coefficients.

We train $\Model$ on multiple NVIDIA A100-SXM2-32GB and A100-SXM4-40GB
GPUs. Due to the use of different teacher/student architectures, the
use of GPUs with different numbers and types, and the share of computational
resources, it is hard to compute exactly the training time of our
method but roughly it took about 1-2 days, 3-4 days, and 4-6 days
to train $\Model$ on CIFAR10/100, TinyImageNet, and ImageNet/Places365/Food101,
respectively.

\begin{table}[H]
\begin{centering}
\resizebox{\textwidth}{!}{%
\par\end{centering}
\begin{centering}
\begin{tabular}{|c|c|c|c|c|c|c|c|c|c|c|c|}
\hline 
\multirow{2}{*}{Dataset} & \multirow{2}{*}{Arch.} & \multicolumn{5}{c|}{Student} & \multicolumn{5}{c|}{Generator}\tabularnewline
\cline{3-12} 
 &  & ${opt}_{\Stu}$ & ${lr}_{\Stu}$ & ${wd}_{\Stu}$ & ${mo}_{\Stu}$ & $n_{\Stu}$ & ${opt}_{\Gen}$ & ${lr}_{\Gen}$ & ${wd}_{\Gen}$ & ${mo}_{\Stu}$ & $n_{\Gen}$\tabularnewline
\hline 
\hline 
\multirow{2}{*}{CIFAR10/100} & $\heartsuit$ & \multirow{3}{*}{SGD} & \multirow{3}{*}{1e-2} & \multirow{3}{*}{5e-4} & \multirow{3}{*}{0.9} & 30 & \multirow{6}{*}{Adam} & \multirow{3}{*}{1e-3} & \multirow{6}{*}{5e-4} & \multirow{6}{*}{-} & \multirow{3}{*}{3}\tabularnewline
\cline{2-2} \cline{7-7} 
 & $\diamondsuit$ &  &  &  &  & 60 &  &  &  &  & \tabularnewline
\cline{1-2} \cline{7-7} 
TinyImageNet & $\heartsuit$ &  &  &  &  & 90 &  &  &  &  & \tabularnewline
\cline{1-7} \cline{9-9} \cline{12-12} 
ImageNet & \multirow{3}{*}{$\clubsuit$} & \multirow{3}{*}{Adam} & \multirow{3}{*}{1e-4} & \multirow{3}{*}{1e-4} & \multirow{3}{*}{-} & \multirow{3}{*}{150} &  & \multirow{3}{*}{1e-4} &  &  & 20\tabularnewline
\cline{1-1} \cline{12-12} 
Places356 &  &  &  &  &  &  &  &  &  &  & 10\tabularnewline
\cline{1-1} \cline{12-12} 
Food101 &  &  &  &  &  &  &  &  &  &  & 5\tabularnewline
\hline 
\end{tabular}}
\par\end{centering}
\caption{Settings of optimizers for student and generator w.r.t. different
datasets and architectures. The teacher/student architecture settings
are ResNet34/ResNet18 ($\heartsuit$), WRN40-2/WRN16-2 ($\diamondsuit$),
and AlexNet/AlexNet ($\clubsuit$). Meanings of abbreviations: $opt$:
optimizer, $lr$: learning rate, $wd$: weight decay, $mo$: momentum,
$n$: number of optimization steps.\label{tab:Opt-Settings-Stu-Gen}}
\end{table}

\begin{table}[H]
\begin{centering}
\resizebox{\textwidth}{!}{%
\par\end{centering}
\begin{centering}
\begin{tabular}{|c|c|c|c|c|c|c|c|c|c|c|c|c|c|}
\hline 
\multirow{2}{*}{Dataset} & \multicolumn{13}{c|}{Training settings}\tabularnewline
\cline{2-14} 
 & $bs$ & $ld$ & $ldep$ & $ep$ & $spe$ & $d_{z}$ & $\alpha$ & $cg$ & $\gamma$ & $pg$ & $pgs$ & $\delta$ & $\nu$\tabularnewline
\hline 
\hline 
CIFAR10/100 & \multirow{2}{*}{256} & \multirow{2}{*}{0.1} & \multirow{2}{*}{100, 200} & \multirow{2}{*}{300} & \multirow{2}{*}{50} & \multirow{5}{*}{256} & \multirow{5}{*}{0.95} & \multirow{2}{*}{No} & - & \multirow{2}{*}{No} & \multirow{2}{*}{-} & \multirow{5}{*}{20} & \multirow{5}{*}{20}\tabularnewline
\cline{1-1} \cline{10-10} 
TinyImageNet &  &  &  &  &  &  &  &  & - &  &  &  & \tabularnewline
\cline{1-6} \cline{9-12} 
ImageNet & \multirow{3}{*}{512} & \multirow{3}{*}{-} & \multirow{3}{*}{-} & \multirow{2}{*}{6000} & \multirow{3}{*}{1} &  &  & \multirow{3}{*}{Yes} & 1.1 & \multirow{3}{*}{Yes} & 200 &  & \tabularnewline
\cline{1-1} \cline{10-10} \cline{12-12} 
Places365 &  &  &  &  &  &  &  &  & 1.1 &  & \multirow{2}{*}{50} &  & \tabularnewline
\cline{1-1} \cline{5-5} \cline{10-10} 
Food101 &  &  &  & 4000 &  &  &  &  & 1.1 &  &  &  & \tabularnewline
\hline 
\end{tabular}}
\par\end{centering}
\caption{Training settings of $\protect\Model$ w.r.t. different datasets.
Meanings of abbreviations: $bs$: batch size, $ld$: learning rate
decay, $ldep$: epochs at which learning rate are decayed, $ep$:
total number of epochs, $spe$: steps per epoch, $d_{z}$: dimensionality
of the noise $z$, $\alpha$: momentum for updating $\tilde{\protect\Gen}$,
$cg$: $\protect\Gen$ is class-conditional or not, $\gamma$: the
scaling hyperparameter in Eq.~\ref{eq:NormReg_loss}, $pg$: Pretraining
$\protect\Gen$ or not, $pgs$: Number of steps for pretraining $\protect\Gen$,
$\delta$: the bound in Eqs.~\ref{eq:LossStu_Practice},\ref{eq:LossGen_Practice},
$\nu$: the bound in Eq.~\ref{eq:LossGen_Practice}.\label{tab:Train-Settings-Stu-Gen}}
\end{table}

\begin{table}[H]
\begin{centering}
\begin{tabular}{|c|c|c|c|c|c|c|c|c|c|}
\hline 
\multirow{2}{*}{Dataset} & \multicolumn{3}{c|}{Student} & \multicolumn{6}{c|}{Generator}\tabularnewline
\cline{2-10} 
 & $\lambda_{0}$ & $\lambda_{1}$ & $\zeta_{0}$ & $\lambda_{2}$ & $\lambda_{3}$ & $\lambda_{4}$ & $\lambda_{5}$ & $\zeta_{1}$ & $\zeta_{2}$\tabularnewline
\hline 
\hline 
CIFAR10/100 & \multirow{5}{*}{1.0} & \multirow{5}{*}{1.0} & \multirow{3}{*}{0.01} & \multirow{5}{*}{1.0} & \multirow{2}{*}{0.0} & \multirow{2}{*}{0.0} & \multirow{2}{*}{1.0} & \multirow{5}{*}{0.1} & \multirow{5}{*}{0.1}\tabularnewline
\cline{1-1} 
TinyImageNet &  &  &  &  &  &  &  &  & \tabularnewline
\cline{1-1} \cline{6-8} 
ImageNet &  &  &  &  & \multirow{3}{*}{0.1} & \multirow{3}{*}{0.1} & \multirow{3}{*}{0.0} &  & \tabularnewline
\cline{1-1} \cline{4-4} 
Places365 &  &  & 0.1 &  &  &  &  &  & \tabularnewline
\cline{1-1} \cline{4-4} 
Food101 &  &  & 0.1 &  &  &  &  &  & \tabularnewline
\hline 
\end{tabular}
\par\end{centering}
\caption{Coefficients of the loss terms in $\protect\Loss_{\protect\Stu}$
(Eq.~\ref{eq:LossStu_Practice}) and $\protect\Loss_{\protect\Gen}$
(Eq.~\ref{eq:LossGen_Practice}).\label{tab:MAD_Coeffs}}
\end{table}

\subsection{Generator Architectures\label{subsec:Generator-Architectures}}

In Table~\ref{tab:Generator-Architectures}, we show different architectures
of the generator w.r.t. different image sizes.

\begin{table}[H]
\begin{centering}
\resizebox{\textwidth}{!}{
\begin{tabular}{cccccccc}
\cline{1-2} \cline{4-5} \cline{7-8} 
Layer & Output size &  & Layer & Output size &  & Layer & Output size\tabularnewline
\cline{1-2} \cline{4-5} \cline{7-8} 
Linear($d_{z}$, 4096) & 4096 &  & Linear($d_{z}$, 16384) & 16384 &  & Linear($d_{z}$, 8192) & 8192\tabularnewline
Reshape (16384, (256, 8, 8)) & 256$\times$4$\times$4 &  & Reshape (16384, (256, 8, 8)) & 256$\times$8$\times$8 &  & Reshape (8192, (512, 4, 4)) & 512$\times$4$\times$4\tabularnewline
ReLU() & 256$\times$4$\times$4 &  & ReLU() & 256$\times$8$\times$8 &  & ResNetBlockY(512, 512) & 512$\times$4$\times$4\tabularnewline
BatchNorm2d(256, 0.1) & 256$\times$4$\times$4 &  & BatchNorm2d(256, 0.1) & 256$\times$8$\times$8 &  & UpsamplingBilinear2d(2) & 512$\times$8$\times$8\tabularnewline
UpsamplingBilinear2d(2) & 256$\times$8$\times$8 &  & UpsamplingBilinear2d(2) & 256$\times$16$\times$16 &  & ResNetBlockY(512, 256) & 256$\times$8$\times$8\tabularnewline
ConvBlockX(256, 128) & 128$\times$8$\times$8 &  & ConvBlockX(256, 128) & 128$\times$16$\times$16 &  & UpsamplingBilinear2d(2) & 256$\times$16$\times$16\tabularnewline
UpsamplingBilinear2d(2) & 128$\times$16$\times$16 &  & UpsamplingBilinear2d(2) & 128$\times$32$\times$32 &  & ResNetBlockY(256, 128) & 128$\times$16$\times$16\tabularnewline
ConvBlockX(128, 64) & 64$\times$16$\times$16 &  & ConvBlockX(128, 64) & 64$\times$32$\times$32 &  & UpsamplingBilinear2d(2) & 128$\times$32$\times$32\tabularnewline
UpsamplingBilinear2d(2) & 64$\times$32$\times$32 &  & UpsamplingBilinear2d(2) & 64$\times$64$\times$64 &  & ResNetBlockY(128, 64) & 64$\times$32$\times$32\tabularnewline
ConvBlockX(64, 32) & 32$\times$32$\times$32 &  & ConvBlockX(64, 32) & 32$\times$64$\times$64 &  & UpsamplingBilinear2d(2) & 64$\times$64$\times$64\tabularnewline
Conv2d(32, 3, 1, 0, 1) & 3$\times$32$\times$32 &  & Conv2d(32, 3, 1, 0, 1) & 3$\times$64$\times$64 &  & ResNetBlockY(64, 32) & 32$\times$64$\times$64\tabularnewline
 &  &  &  &  &  & UpsamplingBilinear2d(2) & 32$\times$128$\times$128\tabularnewline
 &  &  &  &  &  & ResNetBlockY(32, 16) & 16$\times$128$\times$128\tabularnewline
 &  &  &  &  &  & UpsamplingBilinear2d(2) & 16$\times$256$\times$256\tabularnewline
 &  &  &  &  &  & ResNetBlockY(16, 16) & 16$\times$256$\times$256\tabularnewline
 &  &  &  &  &  & Conv2d(32, 3, 3, 1, 1) & 3$\times$256$\times$256\tabularnewline
 &  &  &  &  &  &  & \tabularnewline
\multicolumn{2}{c}{(a) CIFAR10/CIFAR100} &  & \multicolumn{2}{c}{(b) TinyImageNet} &  & \multicolumn{2}{c}{(c) ImageNet/Places365/Food101}\tabularnewline
\end{tabular}}
\par\end{centering}
\caption{Architectures of the generator w.r.t. different datasets. Details
about ConvBlockX and ResNetBlockY are provided in Table~\ref{tab:Generator-Blocks}.\label{tab:Generator-Architectures}}
\end{table}

\begin{table}[H]
\begin{centering}
\begin{tabular}{ccc|c}
\cline{1-1} \cline{3-4} 
ConvBlockX($c_{i}$, $c_{o}$) &  & \multicolumn{2}{c}{ResNetBlockY($c_{i}$, $c_{o}$)}\tabularnewline
\cline{1-1} \cline{3-4} 
Conv2d($c_{i}$, $c_{o}$, 3, 1, 1) &  & \multirow{6}{*}{Conv} & Conv2d($c_{i}$, $c_{o}$, 3, 1, 1)\tabularnewline
ReLU() &  &  & LeakyReLU(0.2)\tabularnewline
BatchNorm2d($c_{o}$, 0.1) &  &  & BatchNorm2d($c_{\ensuremath{i}}$, 0.01)\tabularnewline
Conv2d($c_{i}$, $c_{o}$, 3, 1, 1) &  &  & Conv2d($c_{o}$, $c_{o}$, 3, 1, 1)\tabularnewline
ReLU() &  &  & LeakyReLU(0.2)\tabularnewline
BatchNorm2d($c_{o}$, 0.1) &  &  & BatchNorm2d($c_{o}$, 0.01)\tabularnewline
\cline{3-4} 
 &  & Shortcut & $\begin{cases}
\text{Conv2d(\ensuremath{c_{i}}, \ensuremath{c_{o}}, 1, 0, 1)} & \text{if }c_{i}\neq c_{o}\\
\text{Identity()} & \text{otherwise}
\end{cases}$\tabularnewline
\cline{3-4} 
 &  & Comp. & $y=\text{Conv}(x)+\text{Shortcut}(x)$\tabularnewline
 &  & \multicolumn{2}{c}{}\tabularnewline
(a) &  & \multicolumn{2}{c}{(b)}\tabularnewline
\end{tabular}
\par\end{centering}
\caption{Architectures of ConvBlockX (a) and ResNetBlockY (b).\label{tab:Generator-Blocks}}
\end{table}

\section{Additional Experimental Results}

\subsection{Results of Teacher\label{subsec:Results-of-teacher}}

Table~\ref{tab:Teacher-Acc} reports the results of our teacher on
all the benchmark datasets. On the small datasets, our teacher achieves
very similar performance compared to the one in \cite{fang2021contrastive}.

\begin{table}[H]
\begin{centering}
\resizebox{\textwidth}{!}{%
\par\end{centering}
\begin{centering}
\begin{tabular}{|c|c|c|c|c|c|c|c|c|}
\hline 
\multirow{2}{*}{} & \multicolumn{2}{c|}{CIFAR10} & \multicolumn{2}{c|}{CIFAR100} & TinyIN & ImageNet & Places365 & Food101\tabularnewline
\cline{2-9} 
 & ResNet34 & WRN40-2 & ResNet34 & WRN40-2 & ResNet34 & AlexNet & AlexNet & AlexNet\tabularnewline
\hline 
\hline 
Ours & 95.46 & 94.65 & 78.55 & 75.65 & 66.47 & 56.52 & 50.80 & 65.15\tabularnewline
\hline 
In \cite{fang2021contrastive} & 95.70 & 94.87 & 78.05 & 75.83 & 66.44 & - & - & -\tabularnewline
\hline 
\end{tabular}}
\par\end{centering}
\caption{Classification accuracies of our teacher and of the one in \cite{fang2021contrastive}
on different datasets.\label{tab:Teacher-Acc}}
\end{table}

\subsection{Results of DFKD-Mem with Different Memory Sizes\label{subsec:Results-of-DFKD-Mem}}

\begin{figure}[H]
\begin{centering}
\subfloat[CIFAR100]{\begin{centering}
\includegraphics[width=0.3\textwidth]{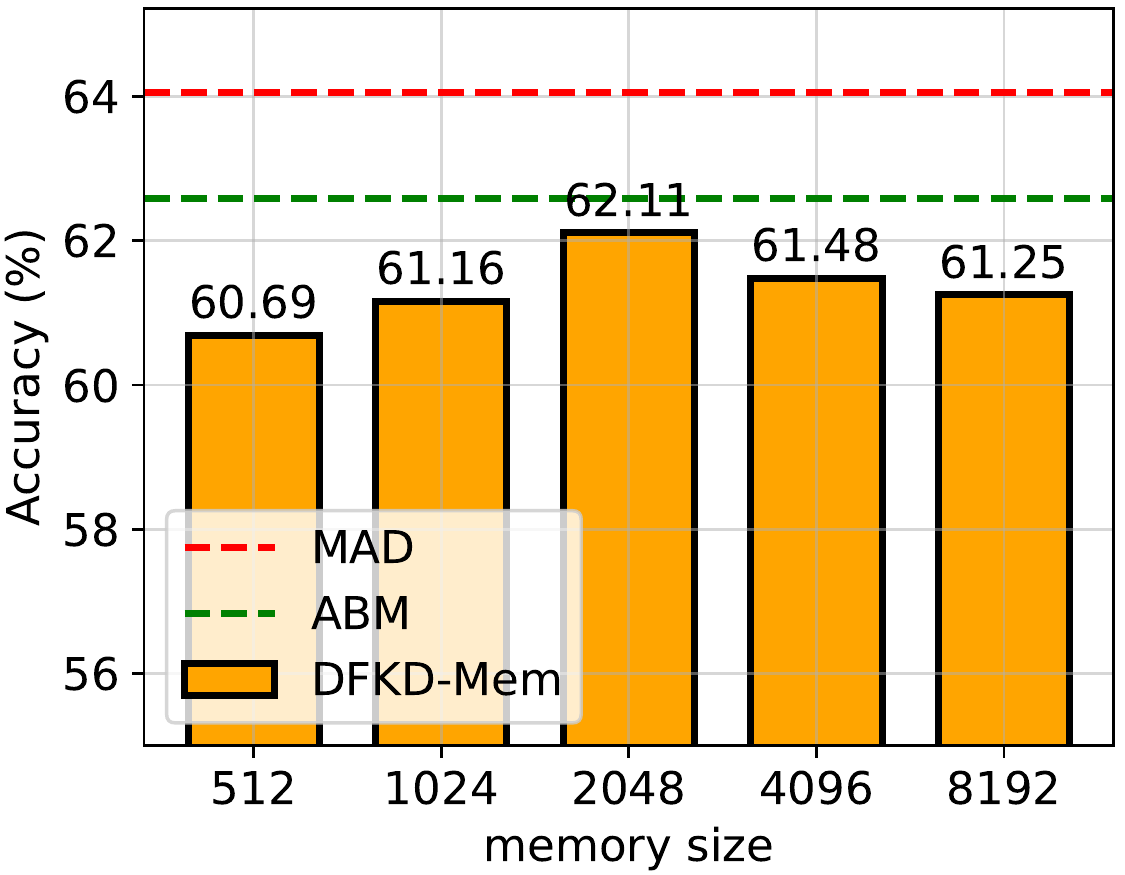}
\par\end{centering}
}\hspace*{0.1\textwidth}\subfloat[ImageNet]{\begin{centering}
\includegraphics[width=0.3\textwidth]{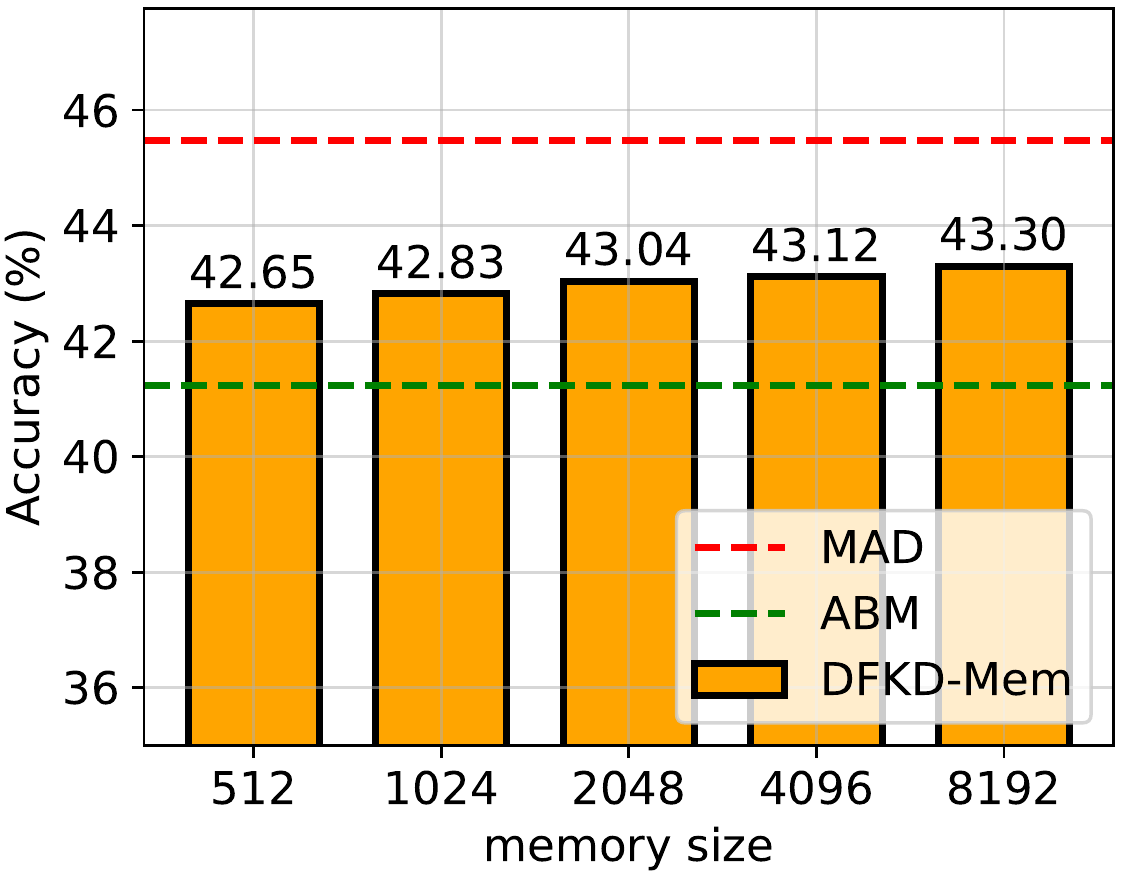}
\par\end{centering}
}
\par\end{centering}
\caption{Classification accuracies of DFKD-Mem with different memory sizes
and of $\protect\Model$, ABM on CIFAR100 and ImageNet.\label{fig:DFKD-Mem-diff-mem-size}}
\end{figure}

In Fig.~\ref{fig:DFKD-Mem-diff-mem-size}, we show the classification
results of DFKD-Mem with different memory sizes on CIFAR100 and ImageNet.
On CIFAR100, DFKD-Mem achieves the best result at memory size = 2048
but still underperforms ABM and MAD. On ImageNet, the performance
of DFKD-Mem is proportional to the memory size and is highest at memory
size = 8192. This result, however, is still worse than that of $\Model$.
Figs.~\ref{fig:KDLoss_on_que},\ref{fig:KDLoss_on_gen} display the
average distillation loss (avg $\Loss_{\text{KD}}$) curves of DFKD-Mem
w.r.t. different memory sizes. We see that increasing the memory size
increases the avg $\Loss_{\text{KD}}$ on memory samples but does
not affect the avg $\Loss_{\text{KD}}$ on samples from $\Gen$. It
is because the avg $\Loss_{\text{KD}}$on memory samples is very small
compared to the counterpart on samples from $\Gen$.

\begin{figure}[H]
\begin{centering}
\subfloat[$\protect\Loss_{\text{KD}}$ on memory samples\label{fig:KDLoss_que_ImageNet}]{\begin{centering}
\includegraphics[width=0.3\textwidth]{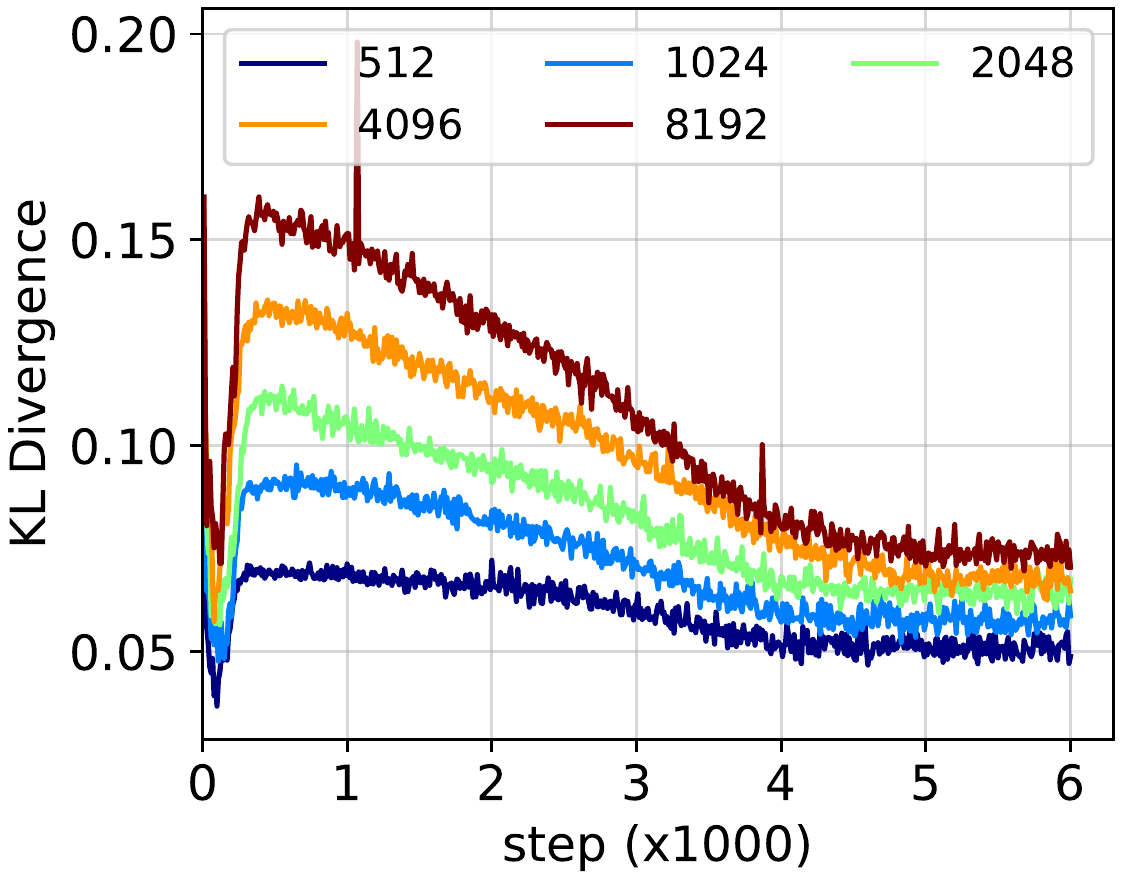}
\par\end{centering}
}\hspace{0.005\textwidth}\subfloat[$\protect\Loss_{\text{KD}}$ on samples from $\protect\Gen$\label{fig:KDLoss_gen_ImageNet}]{\begin{centering}
\includegraphics[width=0.3\textwidth]{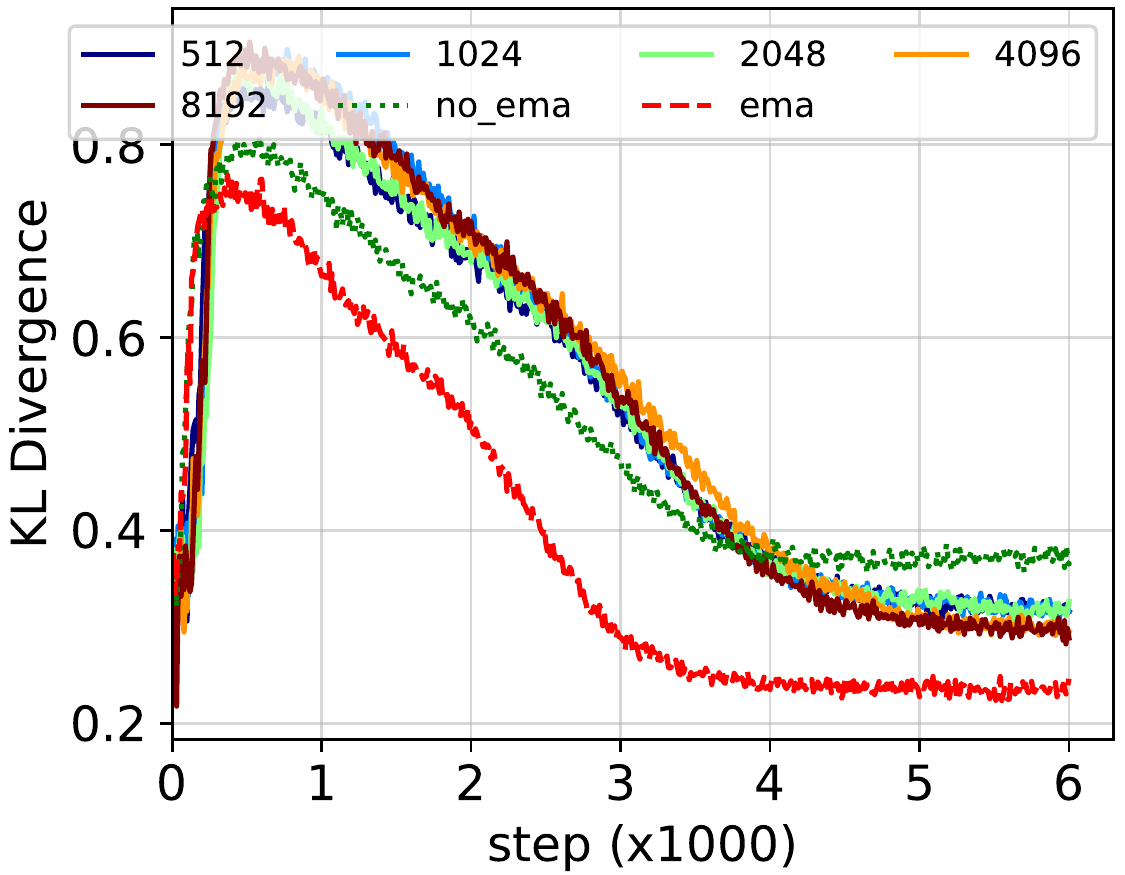}
\par\end{centering}
}\hspace{0.005\textwidth}\subfloat[$\protect\Loss_{\text{KD}}$ on samples from $\tilde{\protect\Gen}$\label{fig:KDLoss_ema_ImageNet}]{\begin{centering}
\includegraphics[width=0.3\textwidth]{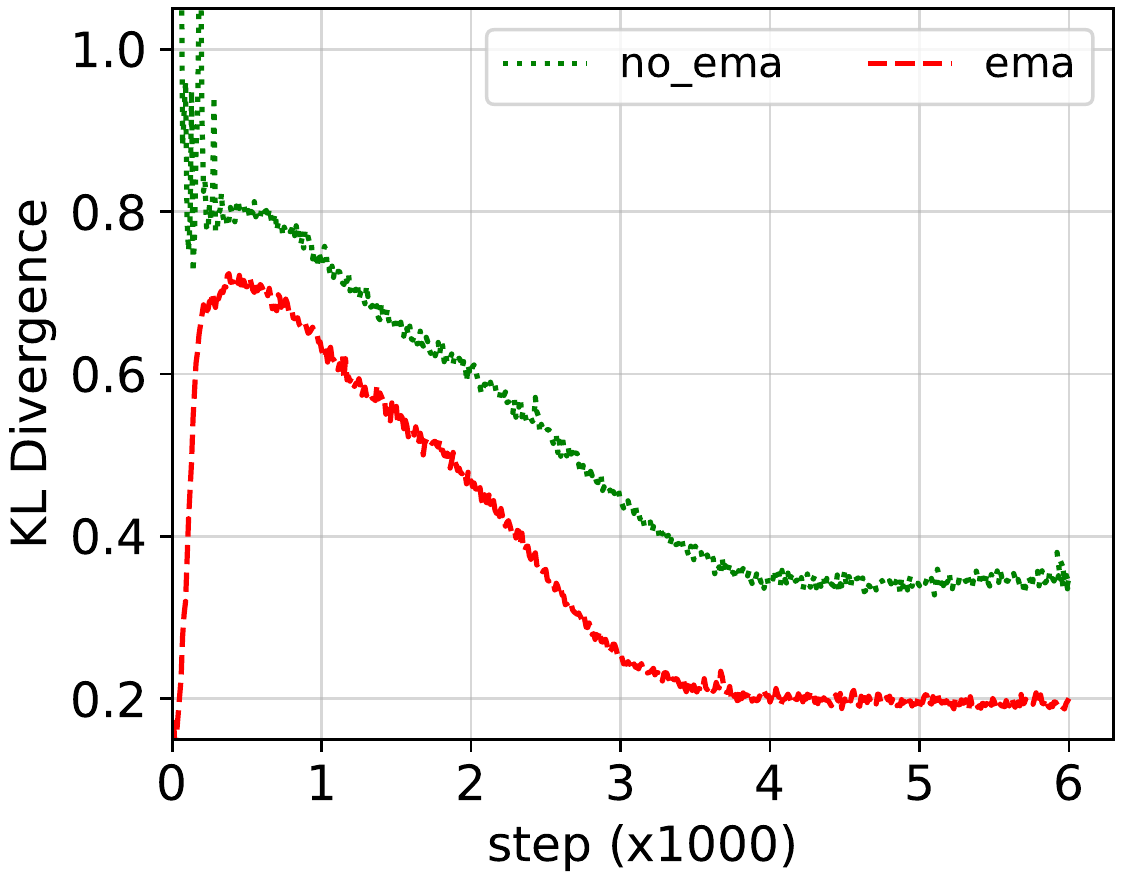}
\par\end{centering}
}
\par\end{centering}
\caption{Distillation loss curves on memory  samples (a) and samples generated
by $\protect\Gen$ (b) and $\tilde{\protect\Gen}$ (c). The numbers
in the legends denote DFKD-Mem with the corresponding memory sizes.
``no\_ema'' and ``ema'' denote ABM and $\protect\Model$, respectively.
The dataset is ImageNet.}
\end{figure}

\subsection{Empirical Analysis of Different Types of Generators\label{subsec:Results-Diff-Cond-Types}}

\begin{figure}
\begin{centering}
\resizebox{\textwidth}{!}{%
\par\end{centering}
\begin{centering}
\begin{tabular}{cccc}
\multicolumn{4}{c}{\includegraphics[width=0.8\textwidth]{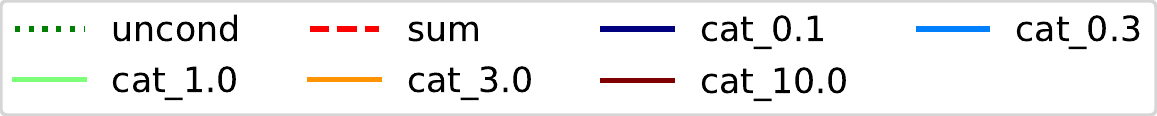}}\tabularnewline
 &  &  & \tabularnewline
\includegraphics[width=0.4\textwidth]{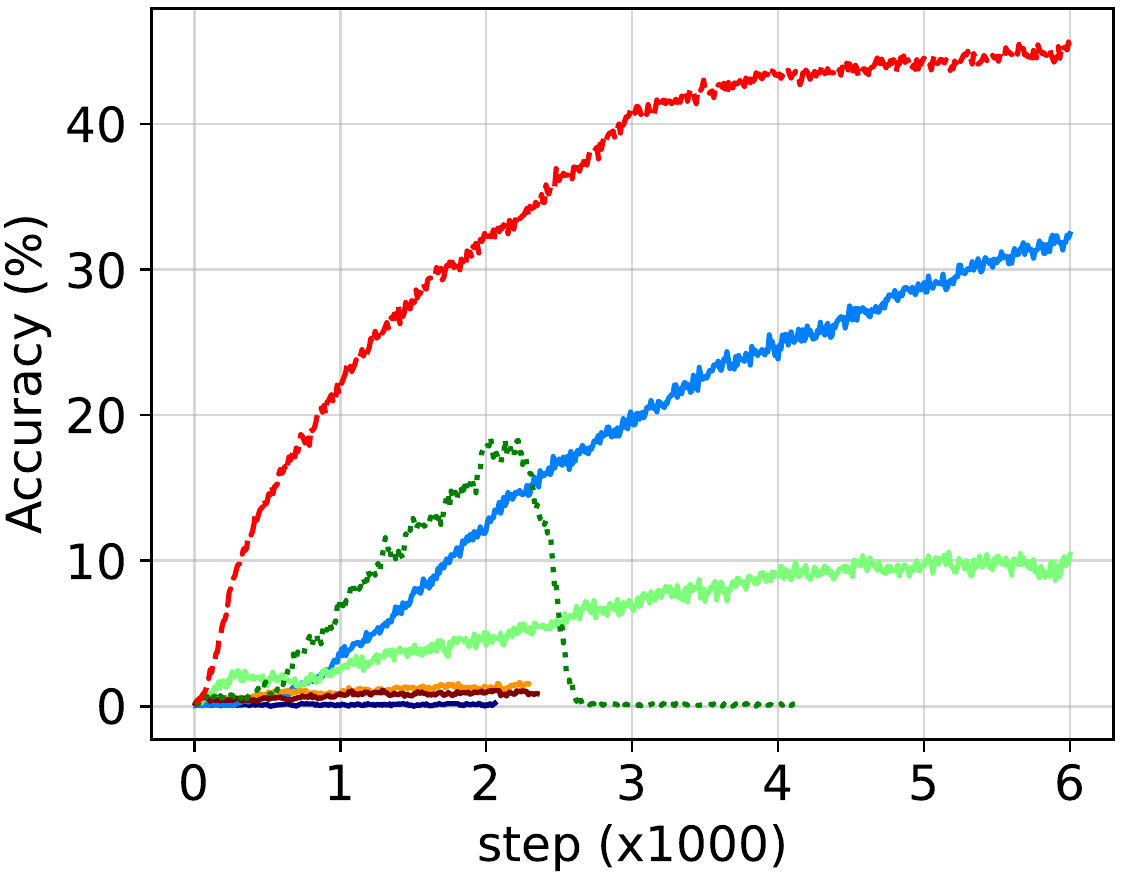} & \includegraphics[width=0.4\textwidth]{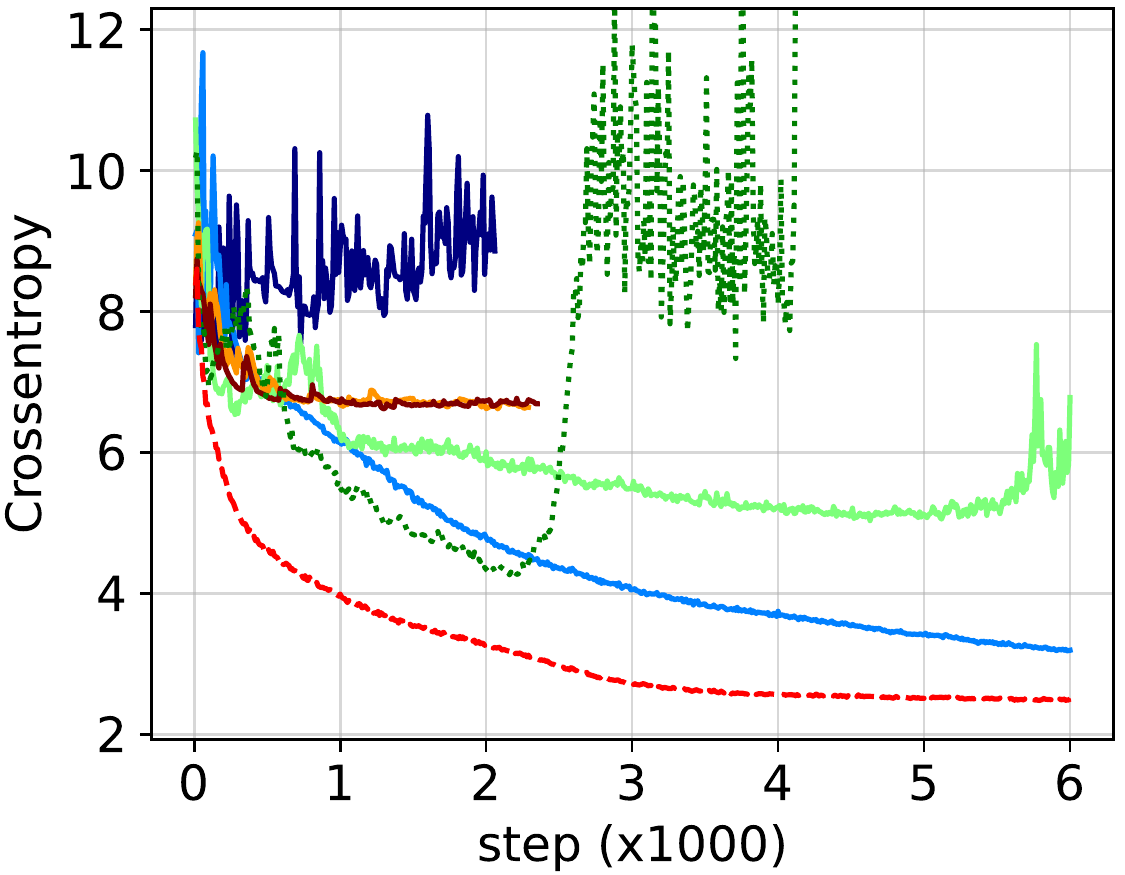} & \includegraphics[width=0.4\textwidth]{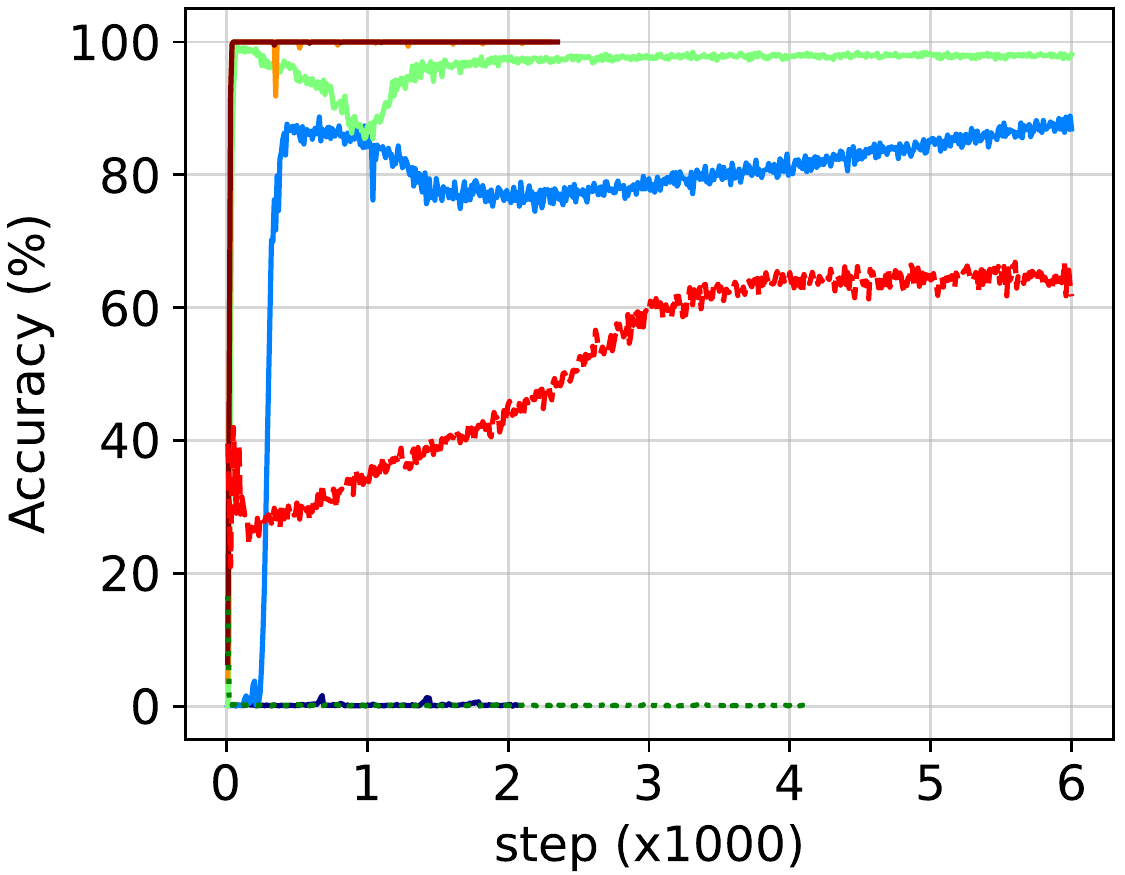} & \includegraphics[width=0.4\textwidth]{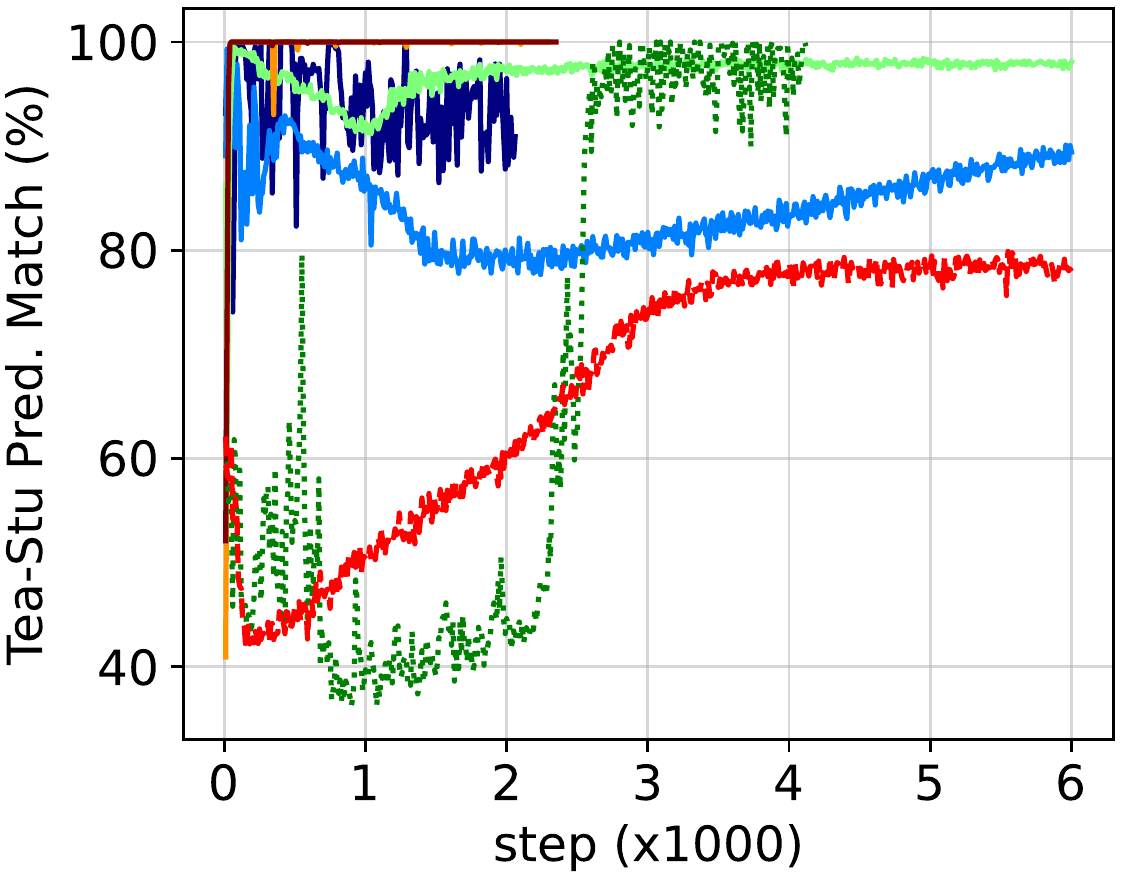}\tabularnewline
(a) Test accuracy & (b) Test crossentropy & (c) Accuracy of $\Tea$  & (d) $\Tea$-$\Stu$ match\tabularnewline
 &  &  & \tabularnewline
\includegraphics[width=0.4\textwidth]{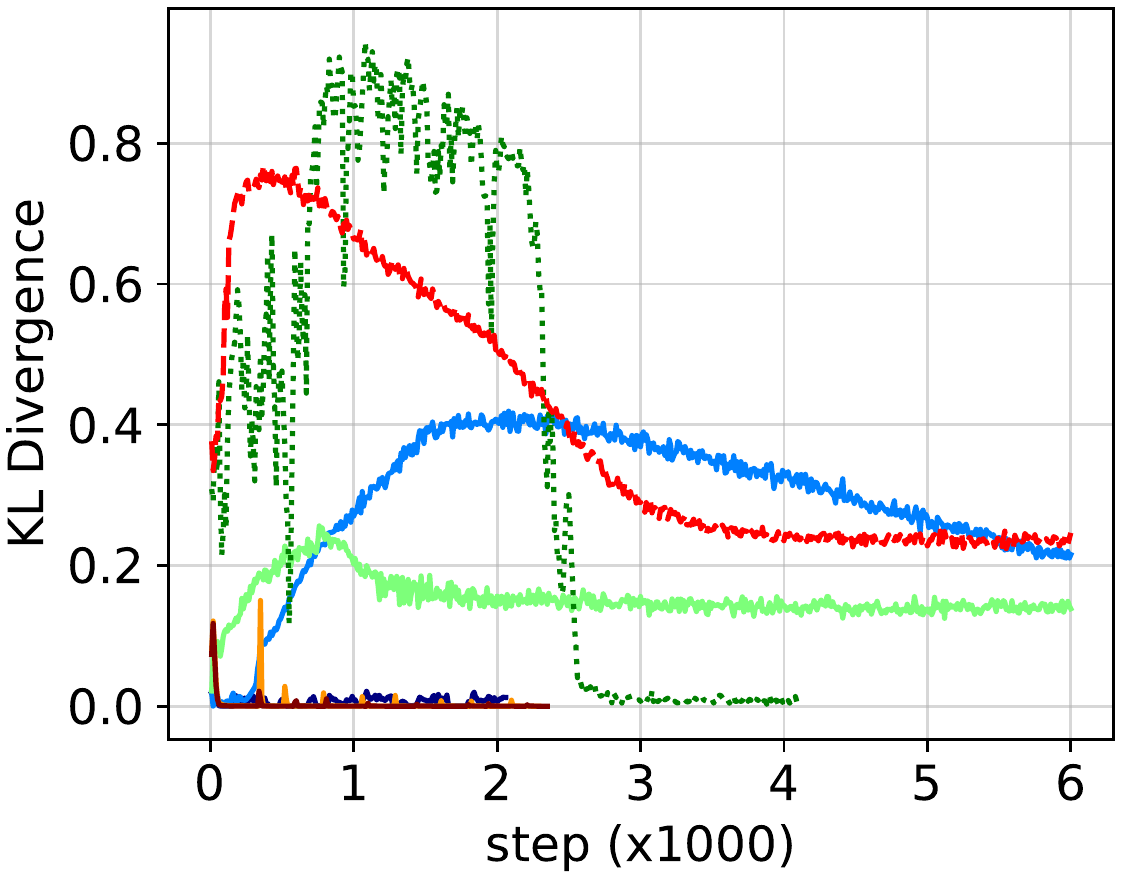} & \includegraphics[width=0.4\textwidth]{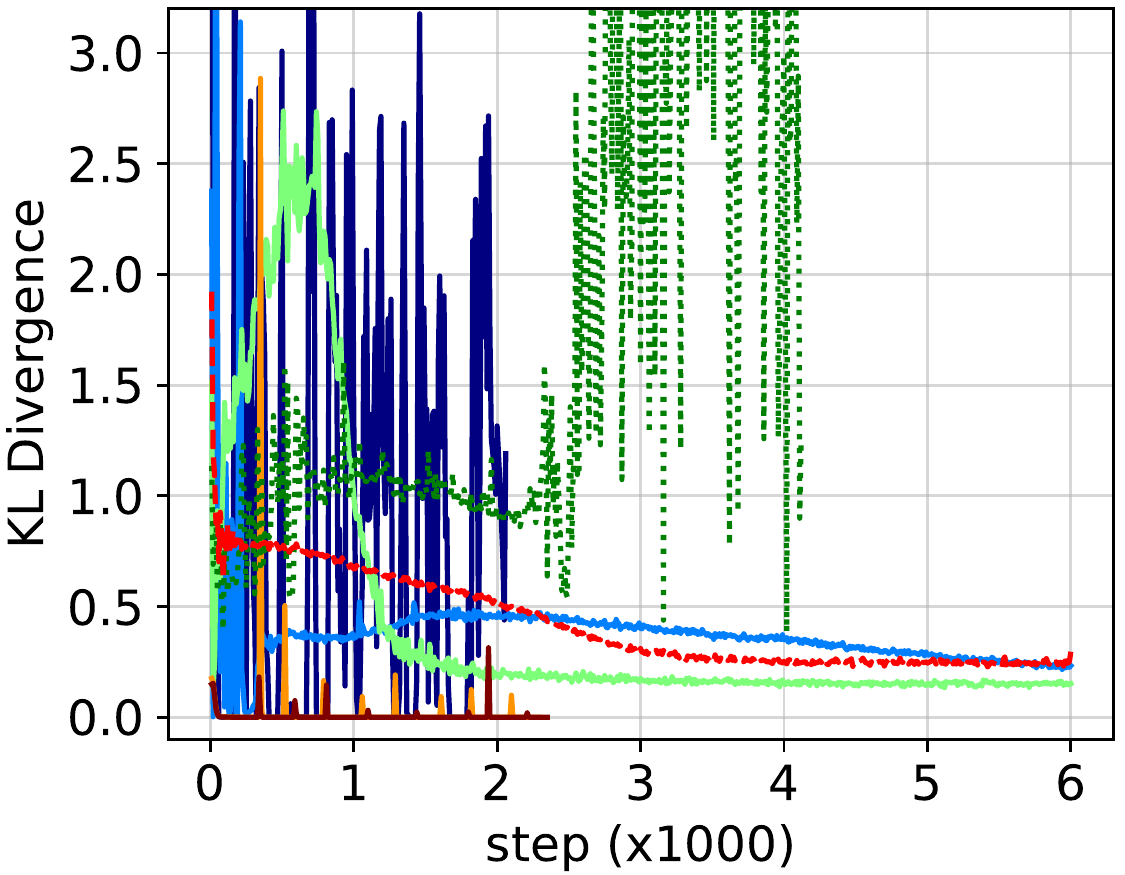} & \includegraphics[width=0.4\textwidth]{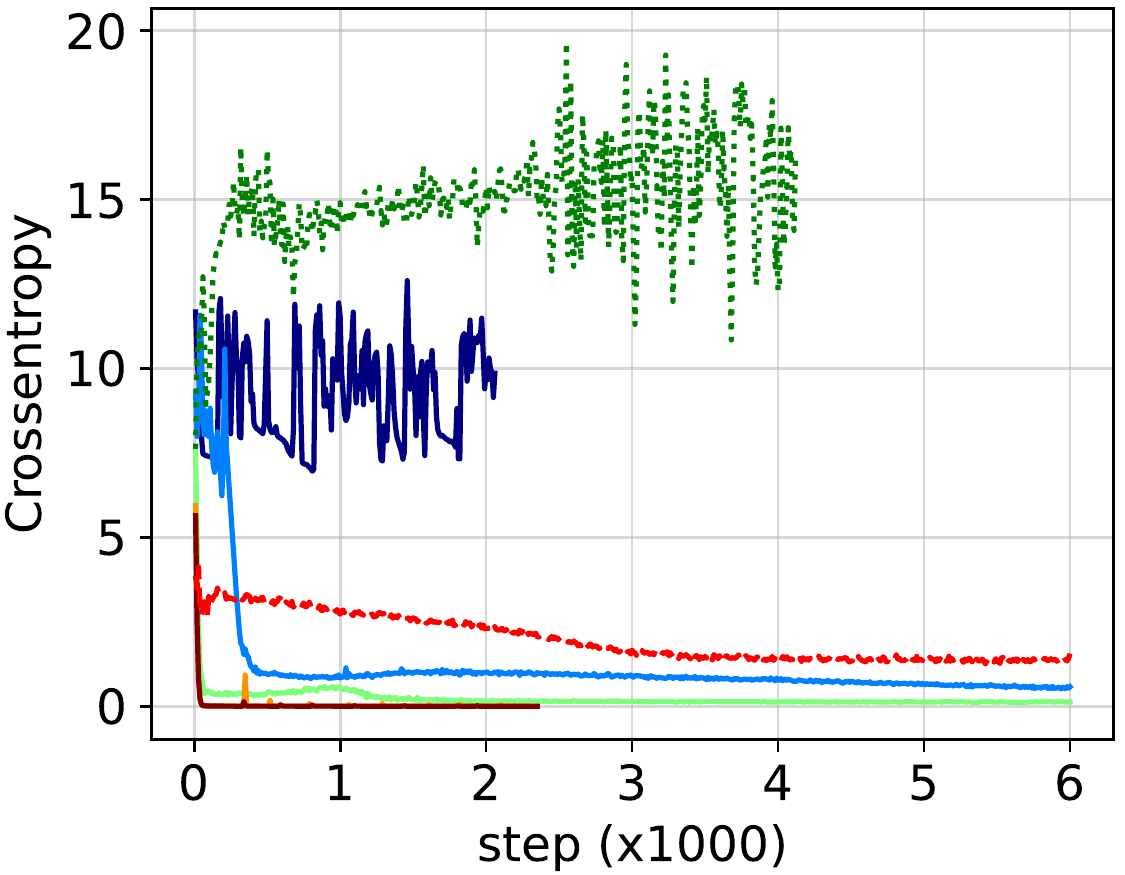} & \tabularnewline
(e) $\Loss_{\text{KD}}$ in training $\Stu$ (Eq.~\ref{eq:MAD_LossS}) & (f) $\Loss_{\text{KD}}$ in training $\Gen$ (Eq.~\ref{eq:ClassCondGen_loss}) & (g) $\Loss_{\text{NLL}}$ in training $\Gen$ (Eq.~\ref{eq:ClassCondGen_loss}) & \tabularnewline
 &  &  & \tabularnewline
\includegraphics[width=0.4\textwidth]{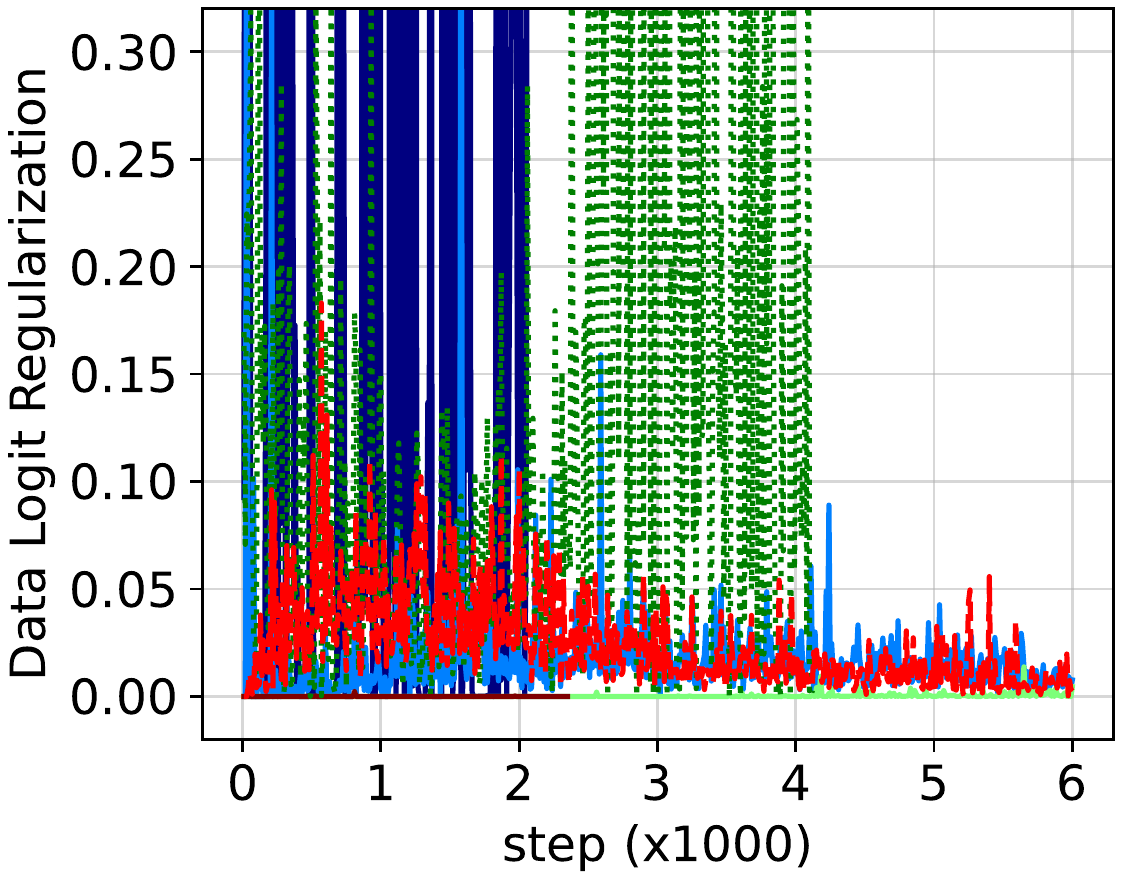} & \includegraphics[width=0.4\textwidth]{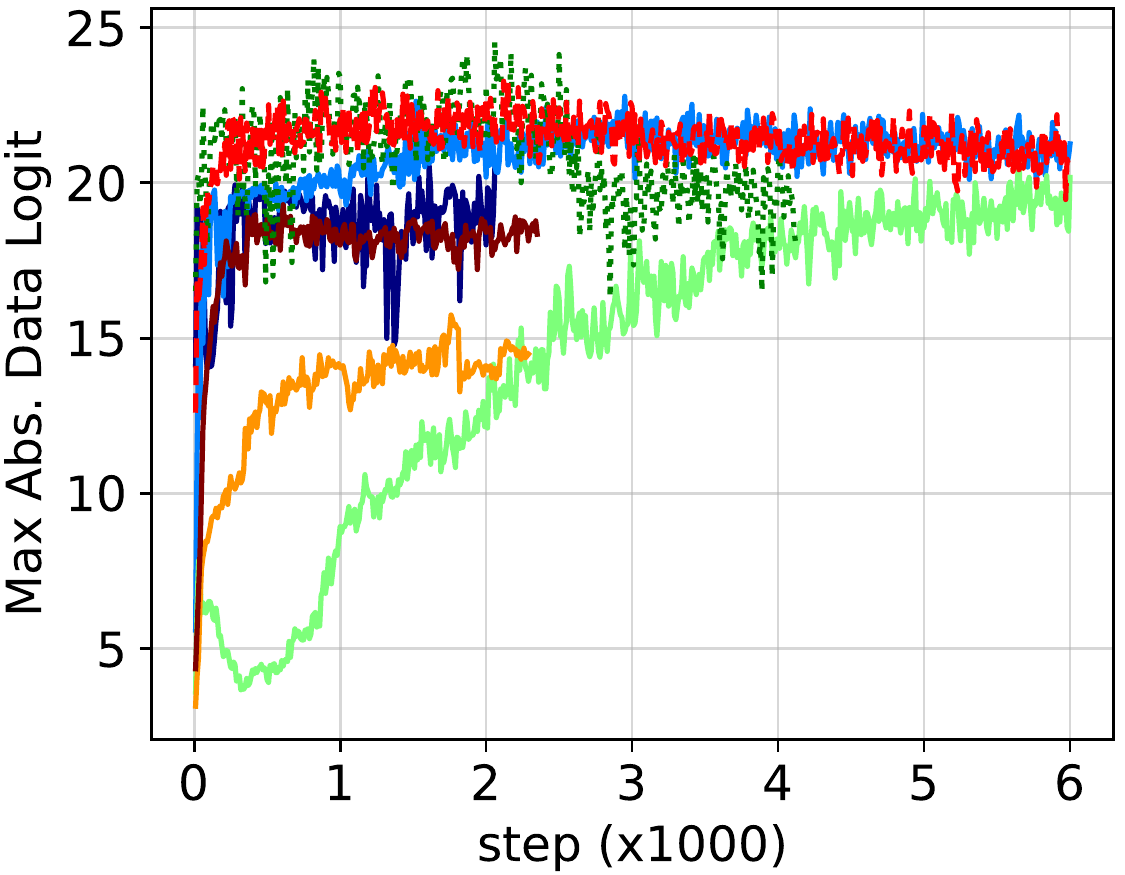} & \includegraphics[width=0.4\textwidth]{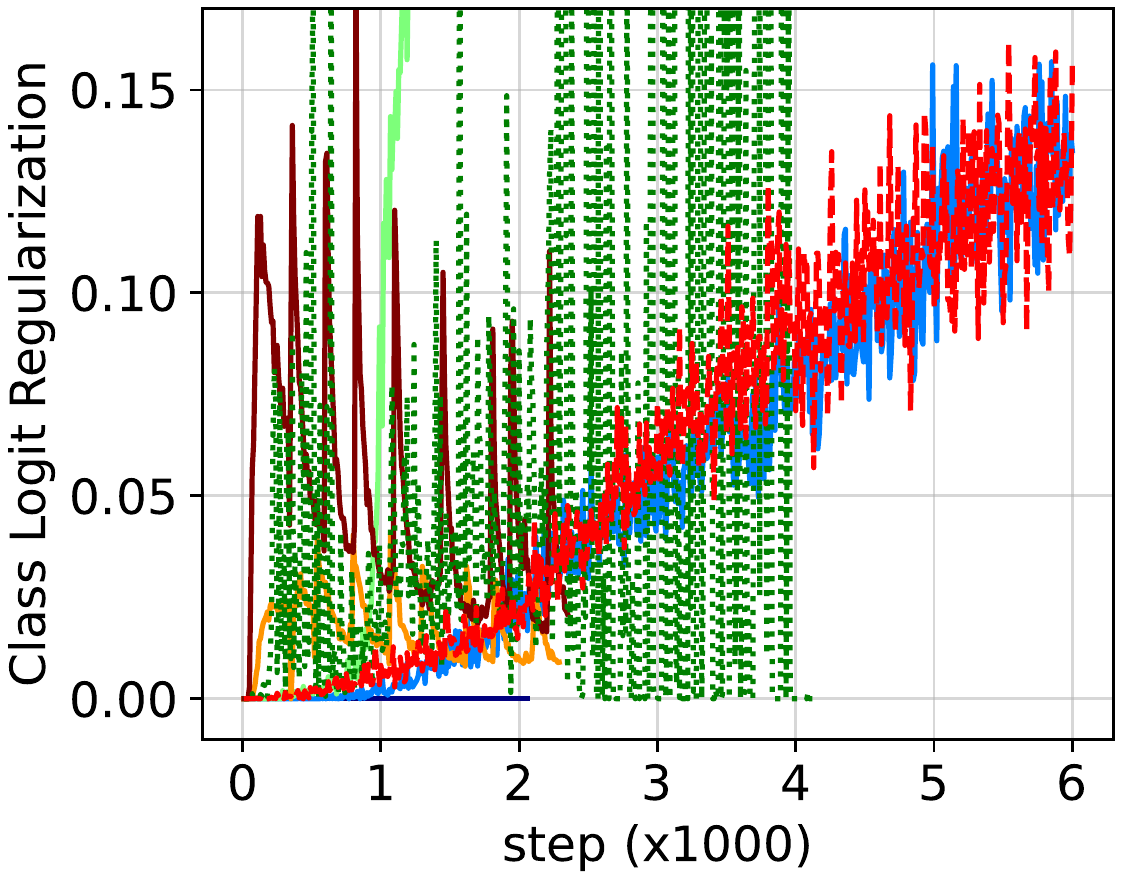} & \includegraphics[width=0.4\textwidth]{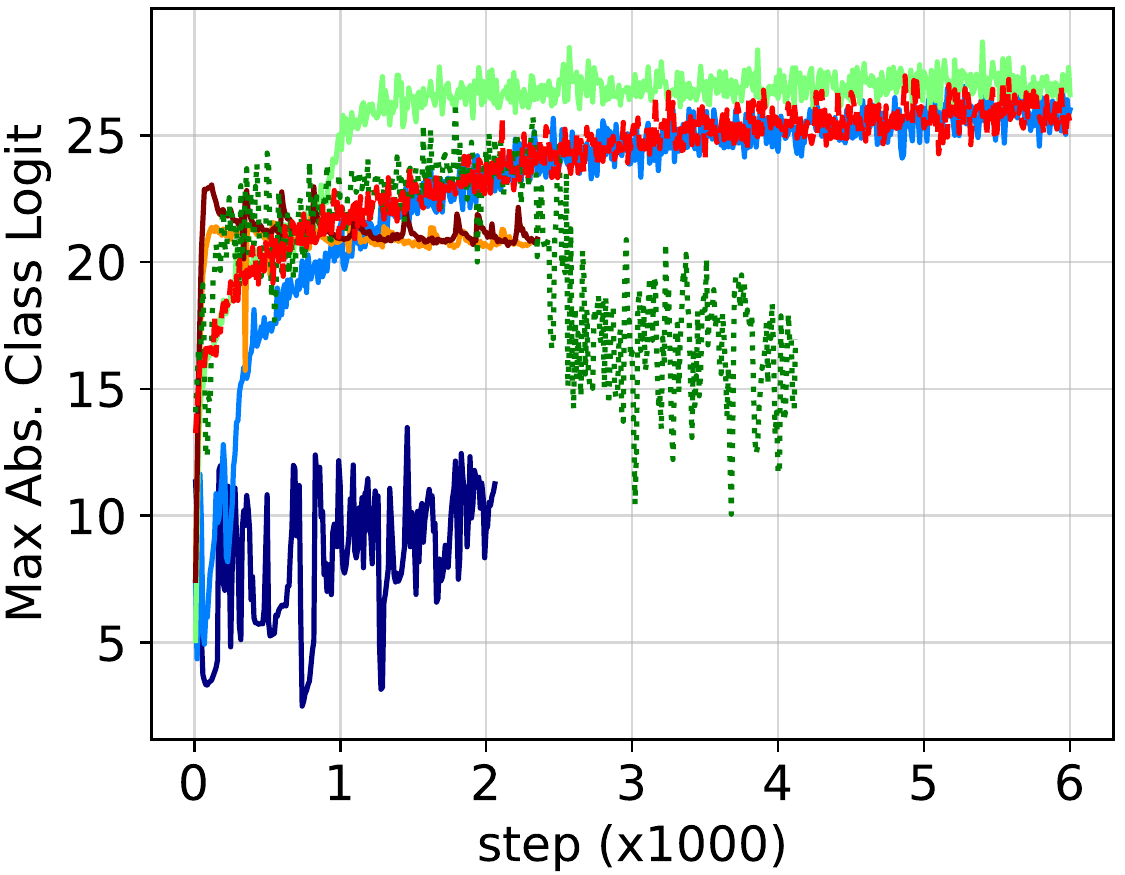}\tabularnewline
(h) $\max(|\Gen_{\text{lg}}|-20,0)$ (Eq.~\ref{eq:LossGen_Practice}) & (i) Avg. of $\max|\Gen_{\text{lg}}|$ over pixels & (j) $\max(|\Tea(\Gen)|-20,0)$ (Eq.~\ref{eq:LossGen_Practice}) & (k) Avg. of $\max|\Tea(\Gen)|$ over classes\tabularnewline
\end{tabular}}
\par\end{centering}
\caption{Various learning curves of $\protect\Model$ with different types
of generators: unconditional (``uncond''), conditional-via-summation
(``sum''), and conditional-via-concatenation (``cat''). For the
``uncond'' generator, $e_{y}$ is set to non-trainable zero vector
and $\lambda_{3}$, $\lambda_{4}$ in Eq.~\ref{eq:ClassCondGen_loss}
are set to 0. For ``cat'' generators, the number behind ``cat''
in the legend indicates the coefficient of $\protect\Loss_{\text{NLL}}$
($\lambda_{3}$) in Eq.~\ref{eq:ClassCondGen_loss}. We tried different
coefficients and found that $\lambda_{3}=0.3$ works best for the
``cat'' generator. Except for Test accuracy and Test crossentropy
which are computed on samples $\protect\Data_{\text{test}}$, all
other quantities are computed on synthetic samples from $\protect\Gen$.\label{fig:Results-Diff-Cond-Types}}
\end{figure}

\begin{figure}
\begin{centering}
\resizebox{\textwidth}{!}{%
\par\end{centering}
\begin{centering}
\begin{tabular}{cccccc}
 & \includegraphics[width=0.3\textwidth]{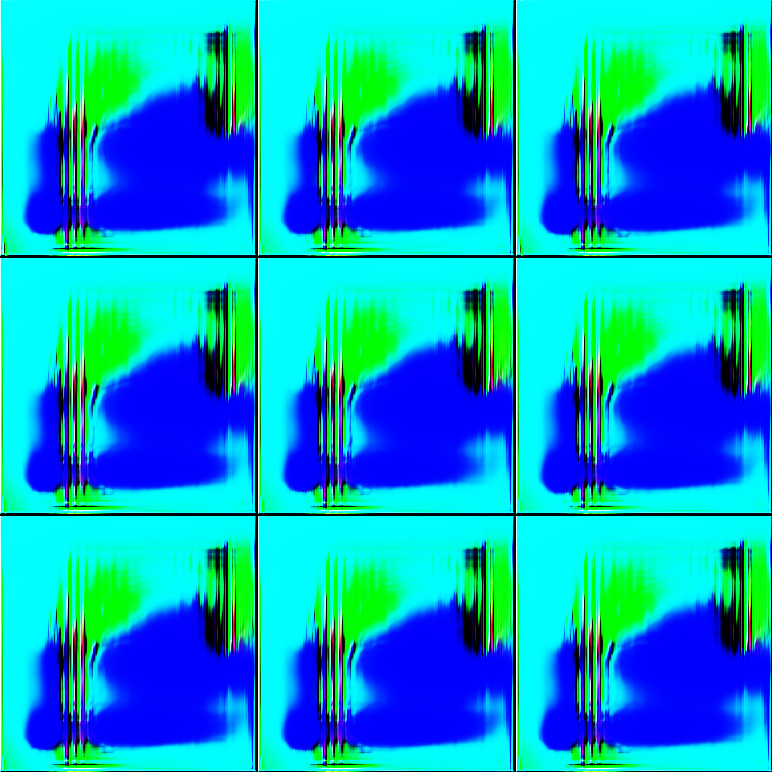} & \includegraphics[width=0.3\textwidth]{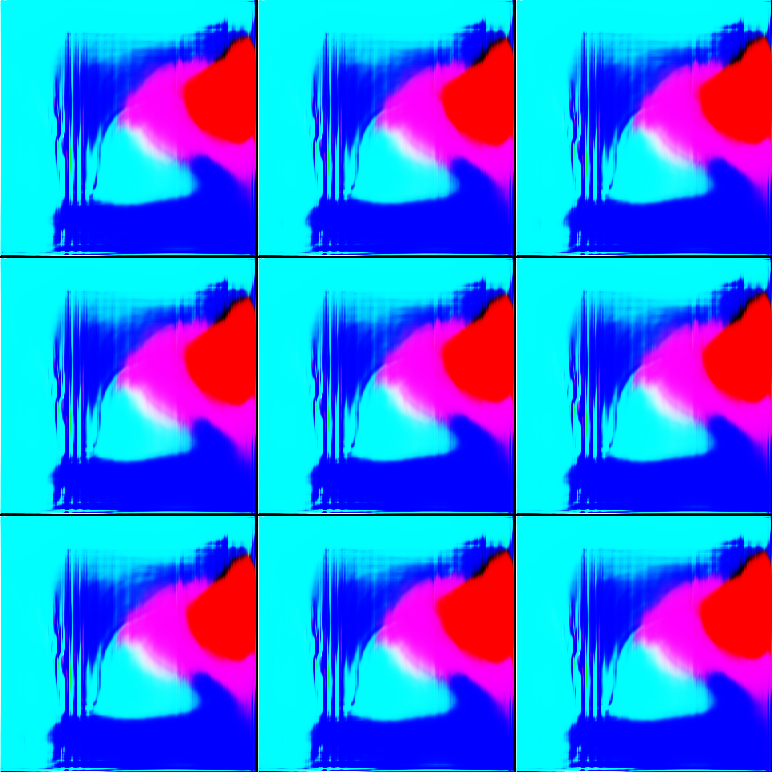} & \includegraphics[width=0.3\textwidth]{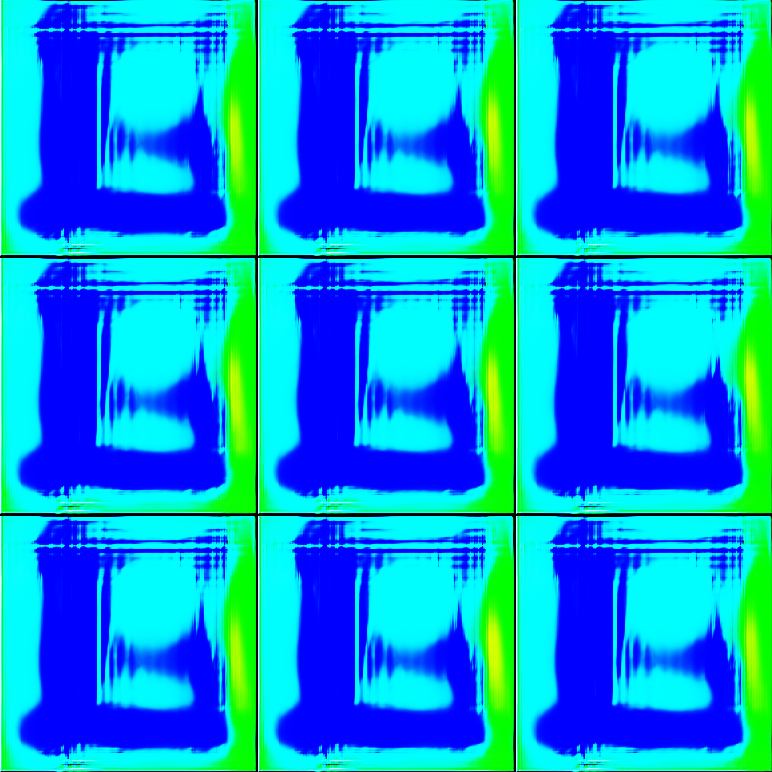} & \includegraphics[width=0.3\textwidth]{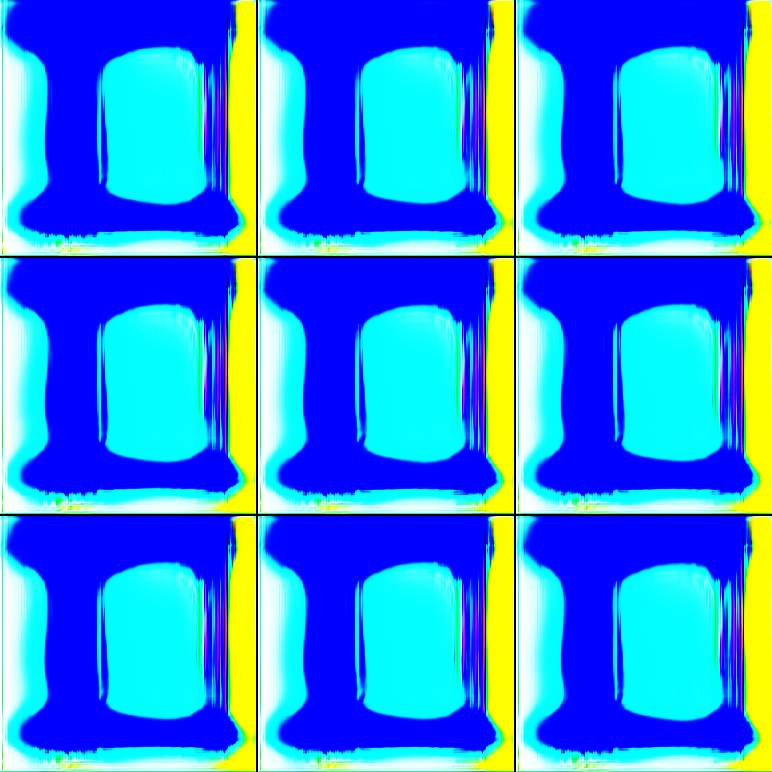} & \includegraphics[width=0.3\textwidth]{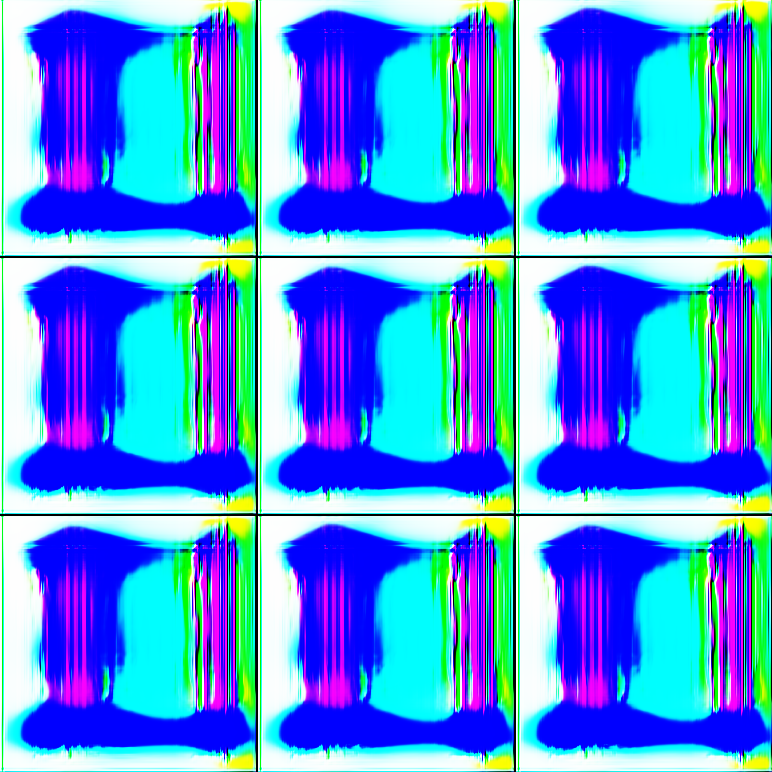}\tabularnewline
Train. Step & 3900 & 3910 & 3920 & 3930 & 3940\tabularnewline
Pred. Class & 107 & 494 & 438 & 711 & 854\tabularnewline
Confidence (\%) & 77.6 & 40.4 & 97.5 & 18.7 & 99.3\tabularnewline
\end{tabular}}
\par\end{centering}
\caption{Visualization of generated samples from the unconditional generator
whose has learning curves shown in Fig.~\ref{fig:Results-Diff-Cond-Types}.
It is obvious that this generator jumps between different spurious
solutions during training, which results in the collapse of the student
in Fig.~\ref{fig:Results-Diff-Cond-Types}a.\label{fig:Visualization-of-UncondGen}}
\end{figure}

In Fig.~\ref{fig:Results-Diff-Cond-Types}, we show the results of
$\Model$ on ImageNet with three different types of generators which
are unconditional (``uncond''), conditional-via-concatenation (``cat''),
conditional-via-summation (``sum''). $\Model$ with the ``uncond''
generator eventually collapses during training but not with the ``cat''
or the ``sum'' generators (Fig.~\ref{fig:Results-Diff-Cond-Types}a).
This is because the ``uncond'' generator has learned to jump between
different spurious solutions as visualized in Fig.~\ref{fig:Visualization-of-UncondGen}.
Among all types of generators, the ``sum'' generator enables stable
training of our model and gives the best accuracy and crossentropy
on $\Data_{\text{test}}$ (Figs.~\ref{fig:Results-Diff-Cond-Types}a,b).
The ``cat'' generator only yields good results at $\lambda_{3}=0.3$
($\lambda_{3}$ is the coefficient of $\Loss_{\text{NLL}}$ in Eq.~\ref{eq:ClassCondGen_loss}).
The reason is that if $\lambda_{3}$ is too small (e.g., 0.1), $\Loss_{\text{NLL}}$
will be high (Fig.~\ref{fig:Results-Diff-Cond-Types}g) and spurious
solutions of $\Gen$ cannot be suppressed. $\Gen$ will jump between
these solutions, leading to high variance when maximizing $\Loss_{\text{KD}}$
(Fig.~\ref{fig:Results-Diff-Cond-Types}f). By contrast, if $\lambda_{3}$
is too big (e.g., 3.0, 10.0), $\Gen$ will be optimized towards predicting
$y$ correctly (small $\Loss_{\text{NLL}}$ as shown in Fig.~\ref{fig:Results-Diff-Cond-Types}g)
rather than generating good adversarial samples for knowledge transfer
from $\Tea$ to $\Stu$ (small $\Loss_{\text{KD}}$ as shown in Fig.~\ref{fig:Results-Diff-Cond-Types}f).
This causes $\Stu$ to achieve tiny $\Loss_{\text{KD}}$ (Fig.~\ref{fig:Results-Diff-Cond-Types}e)
and match $\Tea$ very well (Fig.~\ref{fig:Results-Diff-Cond-Types}d)
on samples from $\Gen$ but generalizes poorly to unseen sample from
$\Data_{\text{test}}$ (Fig.~\ref{fig:Results-Diff-Cond-Types}a).
However, for any value of $\lambda_{3}$, $\Model$ with the ``cat''
generator performs worse than the counterpart with the ``sum'' generator,
and even worse than the counterpart with the ``uncond'' generator
during early epochs of training (Fig.~\ref{fig:Results-Diff-Cond-Types}a).
To explain this phenomenon, we first provide the formulas of the first
layers of the three kinds of generators below as these generators
are only different in the first layer:
\begin{align*}
\text{uncond: } & h=Wz+b\\
\text{cat: } & h=Wz+Ue_{y}+b\\
\text{sum: } & h=Wz+We_{y}+b
\end{align*}
where $W$, $U$, $b$ are trainable weights and bias. We hypothesize
that due to the stochasticity of $z$ sampled from a \emph{fixed}\textbf{
}distribution, $W$ tends to be robust to changes. And since the ``sum''
generator uses $W$ to transform $e_{y}$, its output will not be
affected much by the update of $e_{y}$. In other words, the noise
in updating $e_{\ensuremath{y}}$ is absorbed into the stochasticity
of $z$ via summation. The ``cat'' generator, on the other hand,
uses a different weight matrix $U$ to transform $e_{y}$. Since the
update of $U$ only depends on the current version of $e_{y}$ and
vice versa, and $e_{y}$ can change arbitrarily, updating both $U$
and $e_{y}$ simultaneously in case of the ``cat'' generator can
lead to unstable and nonoptimal\footnote{During the backward pass at step $t$, $U_{t+1}$ is optimal for $e_{y,t}$
and $e_{y,t+1}$ is optimal for $U_{t}$. However, in the forward
pass at step $t+1$, $U_{t+1}$ is used for $e_{y,t+1}$ which leads
to nonoptimality.} training. The ``uncond'' generator does not encounter any problem
with $e_{y}$ like the ``cat'' generator so it can enable $\Model$
to learn faster than the ``cat'' generator.

\section{Derivation of $\protect\Loss_{\text{NormReg}}$ in Section~\ref{subsec:ClassConditionalGen}\label{subsec:Derivation-of-NormRegLoss}}

Recall that in our design of the class-conditional generator, we use
$K$ \emph{trainable} class embedding vectors $e_{1},...,e_{K}$ to
represent $K$ classes in the training data. These embedding vectors
can be regarded as the centers of $K$ Gaussian distributions (or
clusters) $\Normal(e_{k},\mathrm{I})$ ($k=1,...,K$) corresponding
to $K$ classes and are optimized together with the generator $\Gen$
via Eq.~\ref{eq:ClassCondGen_loss}. To prevent these embedding vectors
from changing arbitrarily, we need to constraint their norms to be
smaller than a threshold by minimizing the loss $\Loss_{\text{NormReg}}$
in Eq.~\ref{eq:NormReg_loss}. An important question is ``What is
a reasonable upper bound for the norm of each embedding vector $e_{k}$
?''. 

Let $\xi$ denote the upper bound for the norm of $e_{k}$. By constraining
$\left\Vert e_{k}\right\Vert _{2}$ to be smaller than $\xi$, we
ensure that $e_{k}$ is inside a hyperball of radius $\xi$. Intuitively,
we should choose $\xi$ so that the $K$ Gaussian clusters won't overlap
each other. Note that in high dimensional space, we can generally
treat each Gaussian cluster $\Normal(e_{k},\mathrm{I})$ as a hypersphere
of radius $\sqrt{d_{e}}$ centered at $e_{k}$ ($d_{e}=\dim(e_{k})$).
One simple way to allow these $K$ hyperspheres not to overlap each
other when their centers are inside a hyperball of radius $\xi$ is
to make sure that the total volume of $K$ hyperballs of radius $\sqrt{d_{e}}$
is smaller than the volume of the hyperball of radius $\xi$. Mathematically,
it means:
\begin{align*}
 & K\times\mathcal{V}_{d_{e}}\left(\sqrt{d_{e}}\right)<\mathcal{V}_{d_{e}}\left(\xi\right)\\
\Leftrightarrow & K\times\left(\sqrt{d_{e}}\right)^{d_{e}}\times\mathcal{V}_{d_{e}}\left(1\right)<\xi^{d_{e}}\times\mathcal{V}_{d_{e}}\left(1\right)\\
\Leftrightarrow & K^{1/d_{e}}\sqrt{d_{e}}<\xi
\end{align*}
where $\mathcal{V}_{d}(r)$ denotes the volume of a $d$-ball of radius
$r$. When $d_{e}$ is large, $K^{1/d_{e}}\approx1$ and can be ignored.
Thus, we should choose $\xi$ to have the form $\gamma\times\sqrt{d_{e}}$
with $\gamma\geq1$.

\end{document}